\documentclass[sigconf]{acmart}

\AtBeginDocument{%
  \providecommand\BibTeX{{%
    \normalfont B\kern-0.5em{\scshape i\kern-0.25em b}\kern-0.8em\TeX}}}

\setcopyright{cc} %
\copyrightyear{2023}
\acmYear{2023}
\acmDOI{10.1145/3593013.3593982}

\acmConference[FAccT '23]{2023 ACM Conference on Fairness, Accountability, and Transparency}{June 12--15, 2023}{Chicago, IL, USA}
\acmBooktitle{2023 ACM Conference on Fairness, Accountability, and Transparency (FAccT '23), June 12--15, 2023, Chicago, IL, USA}
\acmPrice{}
\acmISBN{979-8-4007-0192-4/23/06}

\usepackage{numprint}
\usepackage{multirow}

\npthousandsep{,}
\newcommand*{\thead}[1]{\multicolumn{1}{c}{\bfseries #1}}
\newcommand*{\theadline}[1]{\multicolumn{1}{c|}{\bfseries #1}}

\begin{document}

\title[In the Name of Fairness: Assessing the Bias in Clinical Record De-identification]{In the Name of Fairness:\\Assessing the Bias in Clinical Record De-identification}


\author{Yuxin Xiao}
\authornote{Equal contribution.}
\email{yuxin102@mit.edu}
\affiliation{
  \institution{Massachusetts Institute of Technology}
  \country{USA}
}

\author{Shulammite Lim}
\authornotemark[1]
\email{shulim@mit.edu}
\affiliation{
  \institution{Massachusetts Institute of Technology}
  \country{USA}
}

\author{Tom Joseph Pollard}
\email{tpollard@mit.edu}
\affiliation{
  \institution{Massachusetts Institute of Technology}
  \country{USA}
}

\author{Marzyeh Ghassemi}
\email{mghassem@mit.edu}
\affiliation{
  \institution{Massachusetts Institute of Technology}
  \country{USA}
}

\renewcommand{\shortauthors}{Y. Xiao et al.}

\begin{abstract}
Data sharing is crucial for open science and reproducible research, but the legal sharing of clinical data requires the removal of protected health information from electronic health records.
This process, known as de-identification, is often achieved through the use of machine learning algorithms by many commercial and open-source systems.
While these systems have shown compelling results on average, the variation in their performance across different demographic groups has not been thoroughly examined.
In this work, we investigate the bias of de-identification systems on names in clinical notes via a large-scale empirical analysis.
To achieve this, we create 16 name sets that vary along four demographic dimensions: gender, race, name popularity, and the decade of popularity.
We insert these names into 100 manually curated clinical templates and evaluate the performance of nine public and private de-identification methods.
Our findings reveal that there are statistically significant performance gaps along a majority of the demographic dimensions in most methods.
We further illustrate that de-identification quality is affected by polysemy in names, gender context, and clinical note characteristics.
To mitigate the identified gaps, we propose a simple and method-agnostic solution by fine-tuning de-identification methods with clinical context and diverse names.
Overall, it is imperative to address the bias in existing methods immediately so that downstream stakeholders can build high-quality systems to serve all demographic parties fairly.
\end{abstract}



\maketitle

\section{Introduction} \label{sec:1}
The increased availability of clinical datasets \cite{johnson2023mimic, johnson2016mimic, fleurence2014launching} plays a significant role in the recent advancements in machine learning (ML)-aided healthcare systems \cite{shailaja2018machine, ahmad2018interpretable, chen2021ethical, qayyum2020secure}.
In order to share clinical trial data legally, stakeholders must adhere to the Health Insurance Portability and Accountability Act (HIPAA) Safe Harbor provisions by masking 18 types of protected health information (PHI).
If done appropriately, clinical data sharing adds significant value to scientific reproducibility \cite{mcdermott2021reproducibility} at low risk to patient privacy~\cite{seastedt2022global, lehman2021does}.
In this regard, various open-source software \cite{kayaalp2017modes, meystre2010automatic} and commercial companies provide services to de-identify electronic health records (EHRs). 
Named entity recognition (NER) tools \cite{li2020survey, song2021deep, yadav2018survey} in natural language processing (NLP) libraries \cite{honnibal2020spacy, qi2020stanza, akbik2019flair, manning2014stanford} are commonly used in this space. 

Despite the compelling performance of many ML-empowered healthcare systems \cite{topol2019high}, models have been shown to underperform in minorities and minoritized populations, and naive applications can extend and increase existing biases \cite{williams2015racial, ghassemi2020review, chen2021ethical, seyyed2021medical, ghassemi2022medicine, ghassemi2020review, webster2022social}.
Disparities in performance between demographic groups can lead to real harm \cite{zestcott2016examining, marcelin2019impact, hall2015implicit}.
For instance, a state-of-the-art early warning model for acute kidney injury \cite{tomavsev2019clinically} failed to extend to female patients due to its male-dominated training data \cite{cao2022generalizability}.
In de-identification specifically, failing to remove the PHI of certain demographic groups would violate the Safe Harbor regulations. 
This failure could exacerbate the known misuse of data from minorities \cite{browne2015dark, fagan2016stops, ghassemi2022machine} and expose these groups to targeted attacks such as identity theft \cite{anderson2005identity, anderson2006victims}.

\begin{figure*}[t!]
    \centering
    \includegraphics[width=\textwidth]{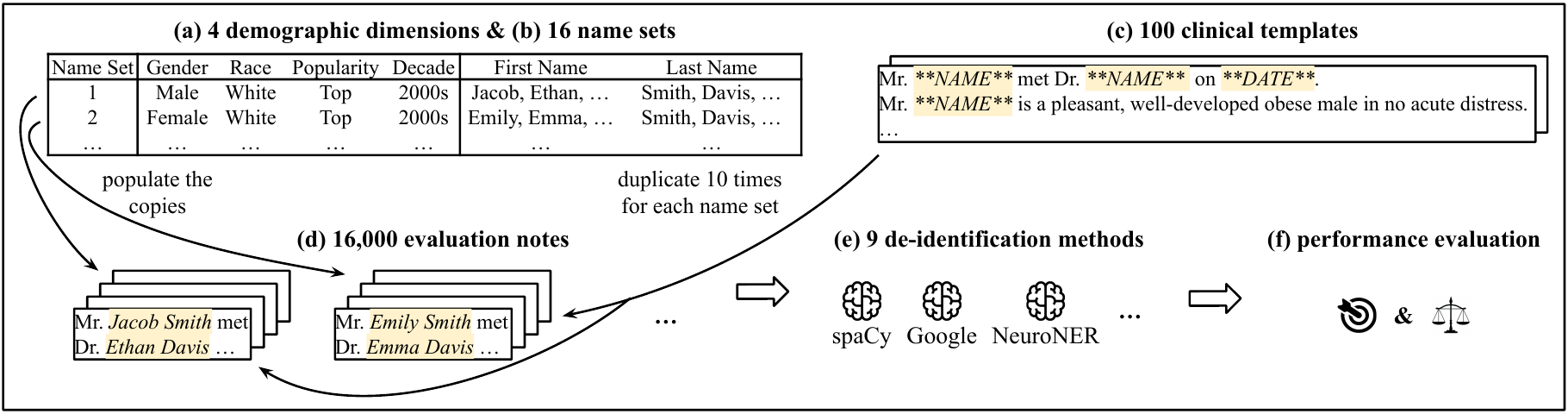}
    \vspace{-17.5pt}
    \caption{Workflow of our empirical study. We identify (a) four demographic dimensions and prepare (b) 16 name sets with diverse settings. For each name set, we duplicate each of the (c) 100 clinical templates ten times and populate the copies with randomly generated names. We then use these (d) \numprint{16000} evaluation notes to assess (e) nine de-identification methods.}
    \label{fig:workflow}
\end{figure*}

In this paper, we audit the performance of de-identification methods on a specific PHI category---names---from clinical notes.
We focus on names because they are correlated with demographic features and are disproportionately identifiable amongst the defined PHI categories.
To date, existing studies \cite{mansfield2022behind, mehrabi2020man, mishra2020assessing} have compared a limited number of baselines on short sentence templates that are much simpler than the real clinical notes.
In contrast, we conduct a large-scale empirical evaluation of nine commercial and open-source de-identification methods based on 16 name sets that vary along four demographic dimensions---gender, race, name popularity, and the decade of popularity---and 100 note templates \cite{lim2023annotated} curated from real-world clinical records.
We adopt the gender and racial categories in the U.S. Social Security \cite{ssaPopularBaby} and Census \cite{censusDecennialCensus} datasets and calculate popularity from name frequency over three selected decades.
While we acknowledge the inherent limitation of using standardized racial categorization and binary gender groups, our work is a first step toward the evaluation of de-identifying names in EHRs, capturing the real harm that gaps could incur.

First, we investigate whether demographic bias exists in clinical de-identification methods.
While some methods attain an overall competitive recall, a majority of the examined methods exhibit statistically significant performance gaps along most demographic dimensions. 
For instance, we note that these methods are, on average, significantly better at recognizing ``rare'' names in White people than ``popular'' names in Asian people.

Second, we assess potential factors contributing to the observed underperformance. 
We find that names with polysemy---other meanings in English---are disproportionately unrecognized, regardless of the associated races. 
Most methods suffer when the gender inferred from a name disagrees with the gender suggested by the semantic context. 
Certain note characteristics, such as length and the number of unique names included, also reduce performance.

Third, we perform fine-tuning on two of the open-source de-identification methods (\texttt{spaCy} \cite{honnibal2020spacy} and \texttt{NeuroNER} \cite{dernoncourt2017neuroner, dernoncourt2016deidentification}) with clinical context and diverse names. 
We find that this significantly improves the methods' overall performance and reduces their demographic bias, especially along the dimensions of race and popularity. 
We advise that this simple, method-agnostic solution should be a minimal first step for practitioners in the de-identification space. 

We contribute a comprehensive analysis of the bias in de-identif-ying names from clinical notes, with insights into the existence of the bias and the cause of the underperformance, and provide a simple mitigation option.
We emphasize that asymmetric de-identification by existing methods could violate legal regulations and is a serious socio-technical ethical issue.
We encourage future work to build upon our results, balancing both de-identification performance and demographic fairness.
\section{Related Work} \label{sec:2}
\paragraph{De-identification}
The HIPAA Safe Harbor regulations require clinical trial data to be properly anonymized before being shared for various purposes \cite{lo2015sharing, tucker2016protecting}.
Toward this goal, the de-identification of EHRs has drawn long-lasting attention from both clinical practitioners and the NLP community \cite{kayaalp2017modes, meystre2010automatic}.
Traditional de-identification methods use rule-based pattern matching \cite{beckwith2006development, friedlin2008software, thomas2002successful, norgeot2020protected} or ML algorithms \cite{uzuner2008identifier, dernoncourt2016deidentification, aberdeen2010mitre, yang2015automatic} for sequence tagging and attain competitive results in the i2b2 (Informatics for Integrating Biology and the Bedside) de-identification challenges \cite{uzuner2007evaluating, stubbs2015annotating}.
Several companies, like Google and Amazon, also provide commercial services to detect and obscure PHI data in plain text.
Along a related line of research, many NLP systems \cite{honnibal2020spacy, qi2020stanza, akbik2019flair, manning2014stanford} can fulfill a similar goal by treating de-identification as an NER problem \cite{li2020survey, song2021deep, yadav2018survey}.

\paragraph{Bias in NLP Systems}
Existing work reports the prevalence of systematic bias in NLP frameworks \cite{shah2020predictive, blodgett2020language}. 
Unfairness in text representations \cite{papakyriakopoulos2020bias, bolukbasi2016man, caliskan2017semantics, kurita2019measuring, zhang2020hurtful} or language models \cite{nadeem2021stereoset, nangia2020crows} can be escalated in downstream applications such as sentiment analysis \cite{bhaskaran2019good, kiritchenko2018examining}, machine translation \cite{savoldi2021gender, stanovsky2019evaluating}, and coreference resolution \cite{rudinger2018gender, zhao2018gender}.
Gender \cite{chaloner2019measuring, sun2019mitigating, maudslay2019s} and racial \cite{blodgett2017racial, davidson2019racial} bias in NLP systems may bring about catastrophic social consequences \cite{hutchinson2020social, sap2019risk}.
In response, researchers have proposed metrics \cite{borkan2019nuanced, czarnowska2021quantifying, jacobs2020meaning} and methods \cite{prost2019debiasing, shin2020neutralizing, huang2020reducing} to mitigate bias in NLP models.

\paragraph{Bias in Healthcare and Other High-Stakes Applications}
Demographic bias exists in healthcare systems \cite{williams2015racial, webster2022social}, typically in an implicit and unconscious way \cite{zestcott2016examining, marcelin2019impact, hall2015implicit}.
For instance, when medical assistance leverages biased artificial intelligence \cite{mehrabi2021survey, hutchinson201950}, the unfairness is usually carried forward to subsequent healthcare practice \cite{ganz2021assessing, gianfrancesco2018potential}.
Hence, addressing the bias here demands joint efforts from both ML researchers \cite{chouldechova2020snapshot, beutel2019putting} and healthcare professionals \cite{byrne2015instilling, ochs2022addressing}. 
Bias could also occur in other high-stakes domains such as job applications \cite{de2019bias, bertrand2004emily, hannak2017bias} and law enforcement \cite{buolamwini2018gender, drozdowski2020demographic, eisenman1995there}.
We leave a detailed discussion of the bias in those areas to future work.

\begin{table*}[t!]
  \centering
  \begin{tabular}{c|cccc|l|l}
    \toprule
        \textbf{Name Set} & \textbf{Gender} & \textbf{Race} & \textbf{Popularity} & \textbf{Decade} & \theadline{\textbf{First Name Examples}} & \thead{\textbf{Last Name Examples}} \\
    \midrule 
    \midrule
        1 & Male & White & Top & 2000s & Jacob, Ethan, Tyler, \dots & Smith, Davis, Brown, \dots \\
        2 & Female & White & Top & 2000s & Emily, Emma, Olivia, \dots & Smith, Davis, Brown, \dots \\
        3 & Male & White & Medium & 2000s & Wade, Ted, Brien, \dots & Waldon, Clapp, Bogle, \dots \\
        4 & Female & White & Medium & 2000s & Mabel, Liz, Terressa, \dots & Waldon, Clapp, Bogle, \dots \\
        5 & Male & White & Bottom & 2000s & Nicki, Leslee, Marti, \dots & Lofft, Lyna, Tamaro, \dots \\
        6 & Female & White & Bottom & 2000s & Glenn, Lyle, Heath, \dots & Lofft, Lyna, Tamaro, \dots \\
    \midrule
        7 & Male & Black & Medium & 2000s & Cedric, Marlon, Ollie, \dots & Booker, Grier, Spikes, \dots \\
        8 & Female & Black & Medium & 2000s & Aisha, Ebony, Jamila, \dots & Booker, Grier, Spikes, \dots \\
        9 & Male & Asian & Medium & 2000s & Zhi, Nguyen, Rajeev, \dots & Ngo, Mao, Ahmed, \dots \\
        10 & Female & Asian & Medium & 2000s & Neha, Priya, Xin, \dots & Ngo, Mao, Ahmed, \dots \\
        11 & Male & Hispanic & Medium & 2000s & Leonel, Camilo, Cruz, \dots & Ceja, Amaro, Recinos, \dots \\
        12 & Female & Hispanic & Medium & 2000s & Celina, Rebeca, Luisa, \dots & Ceja, Amaro, Recinos, \dots \\
    \midrule
        13 & Male & White & Top & 1970s & Patrick, Brian, Eric, \dots & Smith, Davis, Brown, \dots \\
        14 & Female & White & Top & 1970s & Amy, Lisa, Laura, \dots & Smith, Davis, Brown, \dots \\
        15 & Male & White & Top & 1940s & Jerry, George, Frank, \dots & Smith, Davis, Brown, \dots \\
        16 & Female & White & Top & 1940s & Linda, Carol, Nancy, \dots & Smith, Davis, Brown, \dots \\
    \bottomrule 
  \end{tabular}
  \caption{16 name sets of diverse demographic backgrounds and examples of first and last names for each set. Name Sets $1\sim6$ are names with top, medium, and bottom popularity in the 2000s that are also exclusive to the White racial group. Name Sets $7\sim12$ are names with medium popularity in the 2000s that are also exclusive to the Black, Asian, and Hispanic racial groups. Name Sets $13\sim16$ are names with top popularity in the 1970s and 1940s that are also exclusive to the White racial group.}
  \label{tab:name_set} 
  \vspace{-22.5pt}
\end{table*}

\paragraph{Bias in Clinical De-identification}
In light of the discussion above, it is crucial to carefully examine the bias in de-identification methods, given the pivotal role of de-identification in healthcare pipelines.
Previous work \cite{mansfield2022behind, mehrabi2020man, mishra2020assessing} has only compared a small set of baselines based on template sentences that are much simpler than realistic clinical de-identification challenges.
There lacks a holistic analysis that explores the bias in de-identification methods of different categories, the factors leading to the methods' underperformance, and the solution to alleviate the bias.
Therefore, our paper aspires to fill this gap via extensive empirical studies based on 16 name sets with diverse demographic backgrounds, 100 real-world clinical note templates, and nine public and private de-identification methods.
\vspace{-3pt}
\section{Experiment Setup} \label{sec:3}
In this paper, we focus on assessing the bias in de-identifying a specific type of PHI data---people's names---from clinical records.
We choose names amongst the defined PHI types because they are commonly associated with specific demographic features and are particularly identifiable. 

As illustrated in Figure~\ref{fig:workflow}, we first identify (a) four demographic dimensions (i.e., gender, race, name popularity, and the decade of popularity) and prepare (b) 16 name sets with diverse demographic settings in Table~\ref{tab:name_set}.
Each name set consists of 20 first and 20 last names, which can be paired to produce 400 full names in total.
We then curate (c) 100 clinical templates from hospital discharge records \cite{lim2023annotated}.
For each name set, we duplicate each of the 100 templates ten times and fill in full names randomly generated from that name set.
This creates a total of (d) \numprint{16000} notes with \numprint{116160} name mentions for evaluation.
We use these notes to conduct a large-scale empirical analysis of (e) nine de-identification baseline methods to inspect the bias along the four demographic dimensions.\footnote{Our code is available at \url{https://github.com/xiaoyuxin1002/bias_in_deid}.}

\vspace{-3pt}
\subsection{Definition of Demographic Dimensions} \label{sec:3.1}
To measure the demographic information associated with a name, we define the following four demographic dimensions.
\begin{itemize}
    \item The \textbf{gender} of a name refers to the sex assigned at birth to someone with that name, because the phonological property of a name suggests the associated gender \cite{cassidy1999inferring}. We examine two groups for gender: male and female.
    \item The \textbf{race} of a name refers to the expected racial or ethnic identity of someone with that name, reflecting the variation in prevalence that exists between different self-reported racial or ethnic groups \cite{harris2015s}. We consider four racial or ethnic groups: White, Black, Asian, and Hispanic. Other groups are skipped due to prohibitively small community sizes.
    \item The \textbf{popularity} of a name refers to the size of the population of a gender within a decade having that name. We compare three groups here: top, medium, and bottom popularity.
    \item The \textbf{decade} of popularity refers to the decade in which a name is popular in the U.S. in terms of babies being given the name, as name trends change over time \cite{hahn2003drift}. We assess three decade groups: 2000s, 1970s, and 1940s.   
\end{itemize}

\textit{Limitations of Standardized Demographic Categories.} 
We acknowledge the limitation of using standardized self-reported racial categorization and binary gender groups when composing the name sets. 
More fine-grained racial categorizations are possible in future work, and there could be variety in the linguistic norms and naming traditions even within each racial group we consider. 
Transgender and non-binary gender groups are also important to consider in future work, as these groups may use gender-neutral names or have variations in name usage between records.

We use standardized self-reported racial categorization and binary gender groups because it is important to evaluate the performance of de-identification methods on data that is routinely collected in EHRs \cite{bergdall2012cb3}. 
We emphasize that we do not perform any demographic inference as part of a classification system or training set in this work. 
We do not believe that these categories should be viewed as scientific truth and recognize the larger critical interrogation surrounding whether gender and ethnicity should be discerned from names in such systems \cite{lockhart2023name}. 
Instead, we use these categories in the spirit in which they were created by the U.S. Office of Management and Budget to ``monitor and redress social inequality'' \cite{bliss2012race}. 
The examination of the impact of more fluid categorizations of gender, race, and religion is important for future work in this space.

\vspace{-4pt}
\subsection{Construction of Name Sets} \label{sec:3.2}
In this study, we compute the popularity of first names for each gender based on the U.S. Social Security dataset \cite{ssaPopularBaby} across the entire population, rather than for each racial group.
We then select names that are primarily associated with a self-identified racial group with a margin over 10\% based on the mortgage application dataset in \cite{tzioumis2018demographic}.
We note that this is different from picking the most popular names for each racial group independently. 

In the U.S. setting, all top popularity names, as evaluated by absolute frequency ranking, are identified with the White racial group.
For this reason, we consider names associated with the Black, Asian, or Hispanic groups that are of medium popularity.
First names of medium popularity for each race and gender (i.e., Name Sets 3, 4, 7, 8, 9, 10, 11, and 12) are randomly sampled from those with a frequency ranking between 400 and \numprint{8000} in the entire population in the 2000s.
First names of bottom popularity for the White group (i.e., Name Sets 5 and 6) are randomly sampled from those occurring exactly five times in the 2000s.
We set each name set to 20 names since based on the procedure described above, there are only 20 names that are of medium popularity in the 2000s and primarily used by Black males.
We also ensure that first names of top popularity within each gender and decade are mutually exclusive (i.e., no shared first names in Name Sets 1, 2, 13, 14, 15, and 16).

We prepare last names in a similar fashion based on the 2000 Census dataset alone \cite{censusDecennialCensus}, because we assume that the last name popularity is relatively fixed.
Specifically, this means that the most popular last names for the White racial group in the 1970s and 1940s are assigned to be the same as those in the 2000s.

\textit{Limitations of the Datasets.}
We acknowledge that our datasets are limited to the U.S., and therefore, our findings need to be reproduced in other contexts with distinct name distributions.
Furthermore, our use of the mortgage application dataset for self-reported racial matching is limited to those who have the financial security to apply for a loan.
As we do not have access to other sources of names and self-reported races, we use the available data to demonstrate that---even in this presumably more privileged subset of the population---there are de-identification gaps.

\vspace{-4pt}
\subsection{Group Pooling for Demographic Performance Comparisons} \label{sec:3.3}
To evaluate model performance along each demographic dimension, we design experiments that control for other dimensions as follows. 
\begin{itemize}
    \item We assess the impact of \textbf{gender} by pooling and comparing the results of male (i.e., 1, 3, 5, 7, 9, 11, 13, and 15) and female Name Sets (i.e., 2, 4, 6, 8, 10, 12, and 16). Race, popularity, and decade of popularity all vary within these two groups.
    \item We compare performance along \textbf{race} by pooling Name Sets 3 and 4 for the White group, Name Sets 7 and 8 for the Black group, Name Sets 9 and 10 for the Asian group, and Name Sets 11 and 12 for the Hispanic group. These are the male and female names of medium popularity in the 2000s across the four racial groups. 
    \item We examine the influence of \textbf{popularity} by forming and comparing names with varying levels of popularity within the White group, where top popularity is based on Name Sets 1 and 2, medium popularity is based on Name Sets 3 and 4, and bottom popularity is based on Name Sets 5 and 6.    
    \item We evaluate the difference in performance among the three \textbf{decade} groups by comparing the male and female names of top popularity for the White group in each decade: Name Sets 1 and 2 for the 2000s, Name Sets 13 and 14 for the 1970s, and Name Sets 15 and 16 for the 1940s.
\end{itemize}

\vspace{-4pt}
\subsection{Preparation of Clinical Templates} \label{sec:3.4} 
We manually curate 100 clinical note templates based on hospital discharge records from Beth Israel Lahey Health between 2017 and 2019.
We follow the HIPAA Safe Harbor provisions by marking the occurrence of names in the templates and replacing other PHI classes with realistic, synthetic values.
We note that our templates \cite{lim2023annotated} are more complex than those used in existing benchmark datasets \cite{mansfield2022behind, mehrabi2020man, mishra2020assessing}, with an average of \numprint{12893} characters and 3.5 unique names per template and each unique name appearing an average of 2.1 times per template.
This design is more analogous to real-world de-identification challenges and more likely to expose flaws in less effective methods.

\begin{table*}[t!]
  \centering
    \begin{tabular}{c|ccc|cccc}
    \toprule
        \multirow{2}{*}{\textbf{Method}} & \multicolumn{3}{c|}{\textbf{Overall Performance ($\uparrow$)}} & \multicolumn{4}{c}{\textbf{Bias along Dimensions ($\downarrow$)}} \\
        & \textbf{Precision} & \textbf{Recall} & \textbf{F1} & \textbf{Gender} & \textbf{Race} & \textbf{Popularity} & \textbf{Decade} \\
    \midrule 
    \midrule
        \texttt{spaCy} & 0.917$\pm$0.001 & 0.629$\pm$0.001 & 0.746$\pm$0.001 & 0.002$^*\pm$0.001 & 0.013$^*\pm$0.002 & 0.028$^*\pm$0.002 & 0.007$^*\pm$0.002 \\
        \texttt{Stanza} & 0.678$\pm$0.001 & 0.881$\pm$0.001 & 0.766$\pm$0.001 & 0.002$^*\pm$0.001 & 0.016$^*\pm$0.002 & 0.011$^*\pm$0.001 & 0.005$^*\pm$0.001 \\
        \texttt{flair} & 0.920$\pm$0.001 & \textbf{0.974}$\pm$\textbf{0.000} & \textbf{0.946}$\pm$\textbf{0.000} & 0.003$^*\pm$0.000 & \textbf{0.006}$^*\pm$\textbf{0.001} & \textbf{0.008}$^*\pm$\textbf{0.001} & 0.002$^*\pm$0.000 \\
        \texttt{Amazon} & \textbf{0.923}$\pm$\textbf{0.001} & 0.925$\pm$0.001 & 0.924$\pm$0.001 & 0.005$^*\pm$0.001 & 0.022$^*\pm$0.001 & 0.032$^*\pm$0.001 & \textbf{0.001}$\pm$\textbf{0.001} \\
        \texttt{Microsoft} & 0.664$\pm$0.001 & \textbf{0.960}$\pm$\textbf{0.001} & 0.785$\pm$0.001 & 0.003$^*\pm$0.001 & 0.023$^*\pm$0.001 & 0.010$^*\pm$0.001 & 0.006$^*\pm$0.001 \\
        \texttt{Google} & 0.609$\pm$0.001 & 0.869$\pm$0.001 & 0.716$\pm$0.001 & 0.009$^*\pm$0.001 & 0.025$^*\pm$0.001 & 0.014$^*\pm$0.002 & 0.010$^*\pm$0.001 \\
        \texttt{NeuroNER} & \textbf{0.946}$\pm$\textbf{0.001} & 0.944$\pm$0.001 & \textbf{0.945}$\pm$\textbf{0.000} & \textbf{0.001}$\pm$\textbf{0.001} & 0.045$^*\pm$0.001 & 0.026$^*\pm$0.001 & 0.002$\pm$0.001 \\
        \texttt{Philter} & 0.227$\pm$0.001 & 0.794$\pm$0.001 & 0.353$\pm$0.001 & \textbf{0.000}$\pm$\textbf{0.001} & \textbf{0.000}$\pm$\textbf{0.001} & \textbf{0.003}$^*\pm$\textbf{0.002} & \textbf{0.000}$\pm$\textbf{0.001} \\
        \texttt{MIST} & 0.474$\pm$0.001 & 0.751$\pm$0.001 & 0.581$\pm$0.001 & 0.013$^*\pm$0.001 & 0.022$^*\pm$0.002 & 0.017$^*\pm$0.002 & 0.003$^*\pm$0.002 \\
    \bottomrule 
  \end{tabular}
  \caption{Overall performance (higher is better), bias along demographic dimensions (lower is better), and the associated bootstrapped standard error of the examined de-identification methods. We measure the bias with recall equality difference and bold the best two scores in each column. In particular, \texttt{flair} achieves the highest recall and F1 and the lowest bias for race and popularity. Moreover, the asterisk next to a bias score indicates a statistically significant difference in performance at an adjusted significance level ($5\%$ for gender, $0.833\%$ for race, $1.667\%$ for popularity and decade). A majority of the examined methods exhibit statistically significant performance gaps along most demographic dimensions.}
  \label{tab:overall_performance} 
  \vspace{-22.5pt}
\end{table*}

\vspace{-4pt}
\subsection{De-identification Baseline Methods} \label{sec:3.5}
In our large-scale empirical analysis, we examine nine popular de-identification methods of three different categories.
For packages that offer multiple model options, we report the option with the highest performance in our experiments.\footnote{The number of GitHub Stars and citations listed below are accessed on April 24, 2023.}

Three off-the-shelf NLP libraries with the NER function: 
\begin{itemize}
    \item \texttt{spaCy} \cite{honnibal2020spacy} (25.9k GitHub Stars) is widely adopted for industrial information extraction. We choose RoBERTa-base \cite{liu2019roberta}, which is pre-trained on a massive general-purpose corpus, as the backbone of its NER pipeline.
    \item \texttt{Stanza} \cite{qi2020stanza} (6.6k GitHub Stars) is a natural language analysis package. We apply its 18-class NER model variant based on the contextual string representations \cite{akbik2018contextual} and pre-trained on the OntoNotes corpus \cite{weischedel2013ontonotes}. 
    \item \texttt{flair} \cite{akbik2019flair} (12.7k GitHub Stars) is a powerful NLP framework. We employ its large four-class NER model variant built on XLM-R embeddings \cite{conneau2020unsupervised} and document-level features \cite{schweter2020flert} and pre-trained on the CoNLL03 corpus \cite{sang2003introduction}.
\end{itemize}

Three commercial services for PHI detection:
\begin{itemize}
    \item \texttt{Amazon} Comprehend Medical DetectPHI Operation \cite{amazonDetectAmazon} is a HIPAA-eligible NLP service. We segment input notes into pieces shorter than \numprint{20000} characters, the maximum allowed input length, when making the API calls.
    \item \texttt{Microsoft} Azure Cognitive Service for Language PHI Detection \cite{microsoftWhatPersonally} de-identifies PHI information in unstructured texts. We divide notes into slices shorter than \numprint{5120} characters to obey the input length threshold.
    \item \texttt{Google} Cloud Data Loss Prevention De-identification API \cite{googleDeidentifyingSensitive} inspects and redacts sensitive data intelligently. We select the outputs for the class \textit{PERSON\_NAME} and remove the titles before the recognized full names.
\end{itemize}

We note that both \texttt{Amazon} Comprehend Medical DetectPHI Operation and \texttt{Microsoft} Azure Cognitive Service for Language PHI Detection are intended to be used for our specific case of free-text medical note de-identification. \texttt{Google} Cloud Data Loss Prevention De-identification is intended for the general text. 
We use this service because other medically-focused services operated by Google do not operate on free-text notes. 
Specifically, Google Cloud Healthcare API for de-identification \cite{ggHealthcareDeidentify} only operates on FHIR JSON embeddings and DICOM images, and Google Cloud Healthcare Natural Language API \cite{gHealthcareNLP} only recognizes medical knowledge categories.

Three open-source de-identification toolkits:
\begin{itemize}
    \item \texttt{NeuroNER} \cite{dernoncourt2017neuroner, dernoncourt2016deidentification} (212 citations) is an NER tool based on the long short-term memory model \cite{hochreiter1997long}. We use the model pre-trained on the 2014 i2b2 de-identification corpus \cite{stubbs2015annotating} with GloVe word embeddings \cite{pennington2014glove} and collect the outputs for \textit{PATIENT} and \textit{DOCTOR} as the set of recognized names.
    \item \texttt{Philter} (Protected Health Information filter) \cite{norgeot2020protected} (31 citations) leverages the Python NLTK module and regular expressions for rule-based de-identification.
    \item \texttt{MIST} (MITRE Identification Scrubber Toolkit) \cite{aberdeen2010mitre} (156 citations) is a suite of tools for identifying and redacting PHI in free-text medical records. We pre-train the model supplied by the Carafe engine, a conditional random field-based \cite{lafferty2001conditional} sequence tagger, on the 2006 i2b2 de-identification corpus \cite{uzuner2007evaluating} as instructed and view the outputs for the classes \textit{PATIENT} and \textit{DOCTOR} as the set of recognized names.
\end{itemize}

\vspace{-5.5pt}
\subsection{Evaluation of Bias} \label{sec:3.6}
To quantify the bias of each method along each dimension, we follow \cite{mansfield2022behind} by introducing the recall equality difference: the average absolute difference between the recall of each demographic group and that of all the groups along the corresponding demographic dimension. 
More specifically, for dimension $D$ and its entailed set of demographic groups $\mathcal{G}^D = \{\mathcal{G}_1^D, \mathcal{G}_2^D, \dots\}$, recall equality difference $= \frac{1}{|\mathcal{G}^D|} \sum_{\mathcal{G}_i^D \in \mathcal{G}^D} |\text{Recall}(\mathcal{G}_i^D) - \text{Recall}(\mathcal{G}^D)|$.
We use the recall equality difference as the fairness metric since it demonstrates the difference in recall each demographic group would experience while expecting the reported average performance.
We also explore another fairness metric---recall maximum difference---and report the results in Appendix~\ref{app:1.3}.

We carry out the Wilcoxon signed-rank test \cite{woolson2007wilcoxon} for the dimension of gender and the Friedman test \cite{friedman1937use} for the dimensions of race, popularity, and decade to assess the null hypothesis that a de-identification method treats all the groups equally well along a demographic dimension.
After applying the Bonferroni correction \cite{weisstein2004bonferroni}, the adjusted significance levels for gender, race, popularity, and decade are $5\%$, $0.833\%$, $1.667\%$, and $1.667\%$, respectively.
\begin{figure*}[t!]
    \centering
    \includegraphics[width=\textwidth]{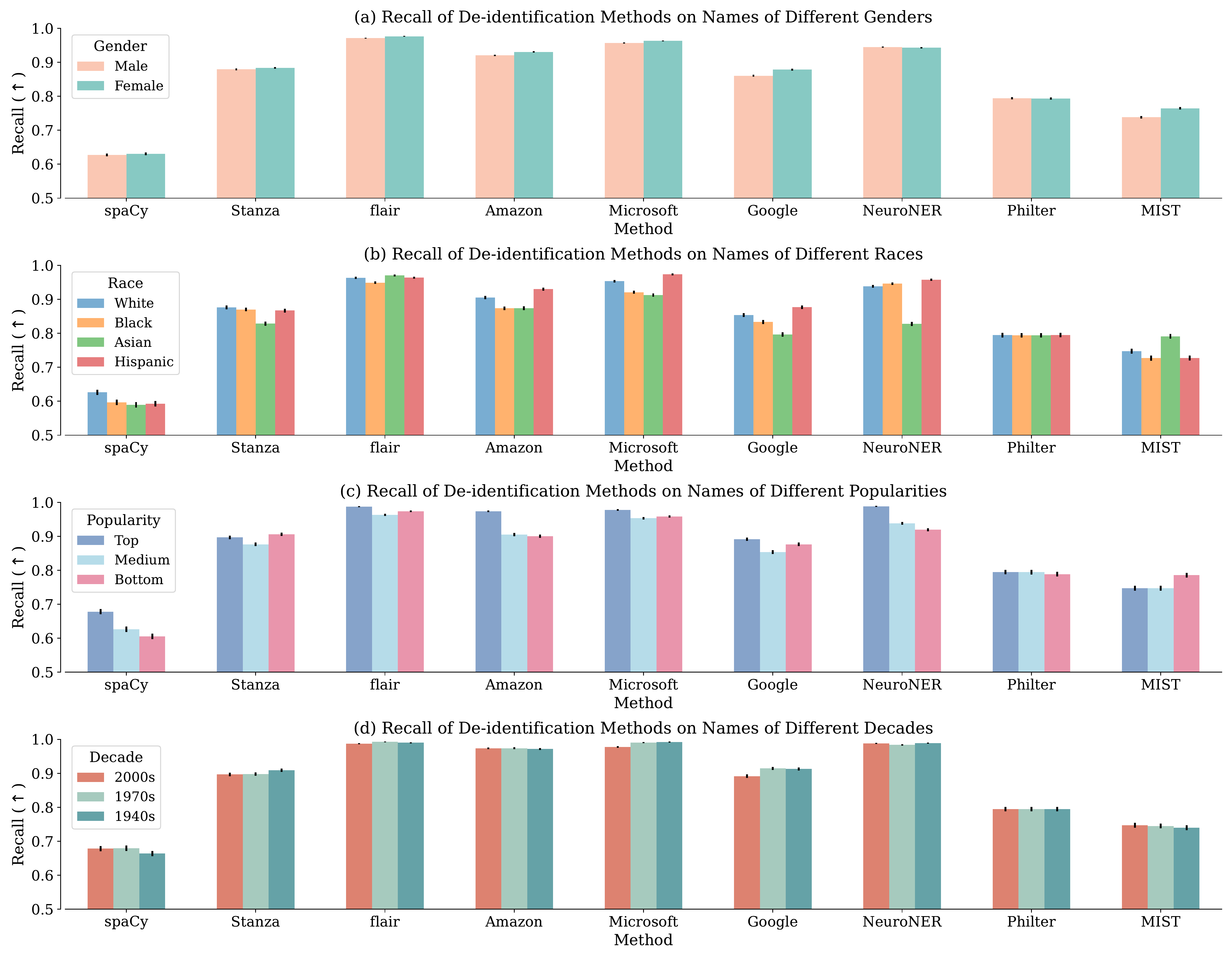}
    \vspace{-24pt}
    \caption{Recall and 95\% bootstrapped confidence interval of the demographic groups along each dimension by each examined de-identification method. Disparities in performance between different groups are more obvious along the dimensions of race and popularity than along the dimensions of gender and decade.}
    \label{fig:dimension_recall}
    \vspace{-12pt}
\end{figure*}

\vspace{-5pt}
\section{Q1: Is There Demographic Bias?} \label{sec:4}
Toward the first question of whether demographic bias exists in de-identification methods, we obtain two key takeaways. 
First, the tested de-identification methods perform differently, with some achieving a relatively high recall. 
Second, a majority of the methods exhibit statistically significant performance gaps along most demographic dimensions.
Such disparities call for urgent review and action to address bias in existing de-identification methods.

\vspace{-5pt}
\subsection{Overall Performance Varies} \label{sec:4.1}
We present the overall performance of the nine de-identification methods in Table~\ref{tab:overall_performance}.
The performance varies across the methods with some methods obtaining a relatively high recall.
In particular, \texttt{flair} performs rather well, especially in recall and F1, probably due to its use of large pre-trained language models and document-level features.
\texttt{NeuroNER} also achieves competitive scores, especially in precision and F1, possibly because it is pre-trained on clinical corpora.
In contrast, \texttt{spaCy} gives the lowest recall, which suggests a high risk of information leakage, albeit its popularity in the NLP community (it has the most GitHub Stars among the three NLP libraries we consider).
Interestingly, \texttt{Google} dramatically underperforms compared to the other two commercial platforms (i.e., \texttt{Amazon} and \texttt{Microsoft}).
As a rule-based method, \texttt{Philter} outputs highly imprecise predictions in the complicated clinical context.

\vspace{-5pt}
\subsection{Significant Demographic Gaps in De-identification Performance} \label{sec:4.2}
We find that a majority of the examined methods demonstrate statistically significant differences in performance along most of the four demographic dimensions. 
Table~\ref{tab:overall_performance} exhibits the recall equality difference and the hypothesis test results, where an asterisk next to a score indicates a statistically significant difference at the corresponding significance level.
In particular, \texttt{Amazon} and \texttt{Google} give the highest recall equality difference for name popularity and the decade of popularity, respectively, which should be a call for action for these commercial services.
Although \texttt{NeuroNER} delivers an overall competitive de-identification performance, its recall equality difference is rather high, especially along the dimensions of race and popularity.
We note that the rule-based \texttt{Philter} has very low bias and that \texttt{flair} achieves not only the highest recall but also relatively low recall equality differences along all four dimensions.

At a fine-grained level, we plot in Figure~\ref{fig:dimension_recall} the recall of the demographic groups along each dimension by each method.
Along the dimension of gender, all the methods score better or equally well for female names than male names.
Nevertheless, these methods act very differently in the four racial groups.
More specifically, \texttt{Stanza} and \texttt{NeuroNER} attain very low recall in the Asian racial group, while \texttt{MIST} scores much higher.
The three commercial services---\texttt{Amazon}, \texttt{Microsoft}, and \texttt{Google}---all perform better in the White and Hispanic racial groups than in the Black and Asian racial groups.
Moreover, the performance of most methods deteriorates with the popularity of names, with the exceptions of \texttt{Stanza} and \texttt{MIST}.
Finally, the disparity in recall among the three groups with different decades of popularity is more moderate.
\texttt{Stanza}, \texttt{Microsoft}, and \texttt{Google} are more capable of recognizing popular names from more recent decades, while \texttt{spaCy} behaves oppositely.

\begin{figure*}[t!]
    \centering
    \includegraphics[width=\textwidth]{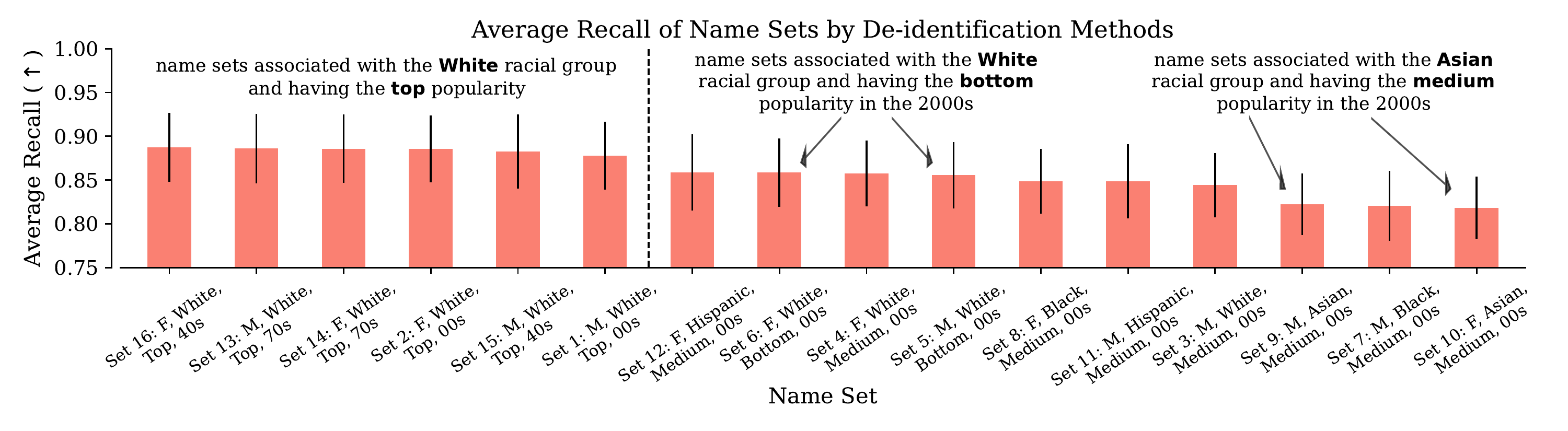}
    \vspace{-26pt}
    \caption{Average recall and standard error of each name set by the examined de-identification methods, ordered by decreasing recall. The average recall on name sets with top popularity exceeds the other sets by a clear margin. Moreover, the methods are, on average, more capable of recognizing less popular names associated with the White racial group compared to more popular names associated with the Asian racial group.}
    \label{fig:set_recall}
    \vspace{-8pt}
\end{figure*}

We visualize in Figure~\ref{fig:set_recall} the recall of the 16 name sets averaged across the examined methods to further examine performance disparities.
We observe that the average recall of the name sets with top popularity (i.e., Name Sets 1, 2, 13, 14, 15, and 16) outperforms the other sets.
In addition, we note that the least popular names associated with the White racial group (i.e., Name Sets 5 and 6) score higher on average recall than the more popular names associated with the Asian racial group (i.e., Name Sets 9 and 10).

\begin{figure*}[t!]
    \centering
    \includegraphics[width=\textwidth]{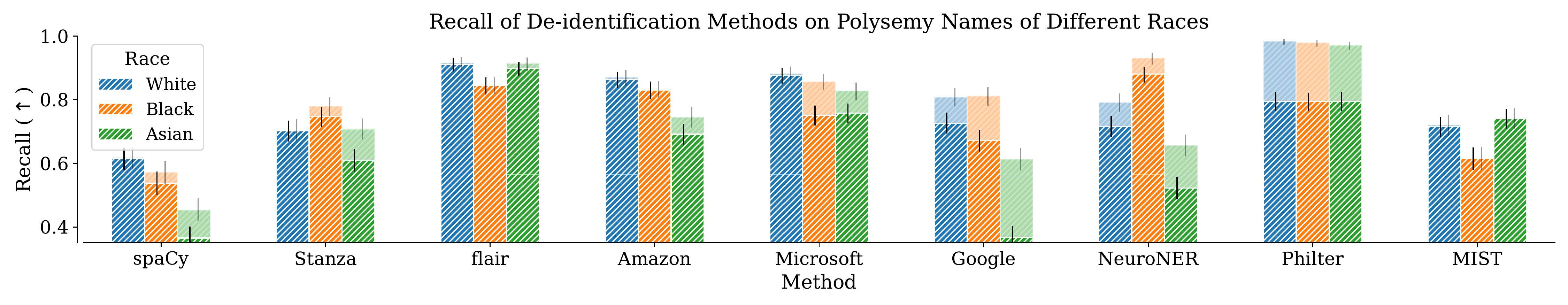}
    \vspace{-23pt}
    \caption{Recall and 95\% bootstrapped confidence interval on polysemy first names associated with three racial groups by each examined de-identification method. The recall ranking among the three groups remains relatively consistent for most methods as that based on the original setting in Figure~\ref{fig:dimension_recall} (b). The increase in recall illustrated by the lighter color bar refers to the partially correct de-identification of non-polysemy last names.}
    \label{fig:polysemy_recall}
    \vspace{-8pt}
\end{figure*}

\section{Q2: What Leads to De-Identification Underperformance?} \label{sec:5}
For the second question of what factors contribute to the underperformance, we draw three critical findings.
\begin{itemize}
    \item Polysemy names account for methods' underperformance but not necessarily their demographic bias.
    \item Most methods are better at recognizing names in agreement with the gender suggested by the local context.
    \item Longer templates with more unique names and medication injection histories make de-identification harder.
\end{itemize}

\vspace{-6pt}
\subsection{Polysemy Names Cause Underperformance} \label{sec:5.1}
To understand what names are the hardest to recognize, we calculate the recall of each sampled name.
We observe that names with the lowest recall usually have other meanings in English (i.e., polysemy).
For instance, ``\textit{An}'' in \textit{An Dizon} and \textit{An Son}---the two names with the lowest recall---is both a prevalent determiner in English and a first name of medium popularity associated with the Asian female group.
``\textit{Cleveland}'' in \textit{Cleveland Spikes}---the fifth hardest name by recall---is both a large city in the U.S. and a first name of medium popularity for the Black male group.

Therefore, we prepare five polysemy first names for each of the White, Black, and Asian racial groups as follows:
\begin{itemize}
    \item White: Sydney, Faith, Forest, Cliff, June
    \item Black: Quincy, Cleveland, Kenya, Prince, Ivory
    \item Asian: Asian, Thai, King, Long, Young, Can
\end{itemize}
These sets share the same gender, name popularity, and the decade of popularity and only differ in race.
Since we can only find five polysemy first names from the Black racial group and not enough polysemy first names from the Hispanic group that meet this requirement, we limit all sets to five names and omit the Hispanic group here.
We then follow the procedure in Sec~\ref{sec:3} and evaluate the methods on the polysemy first names listed above.

\begin{figure*}[t!]
    \centering
    \includegraphics[width=0.75\textwidth]{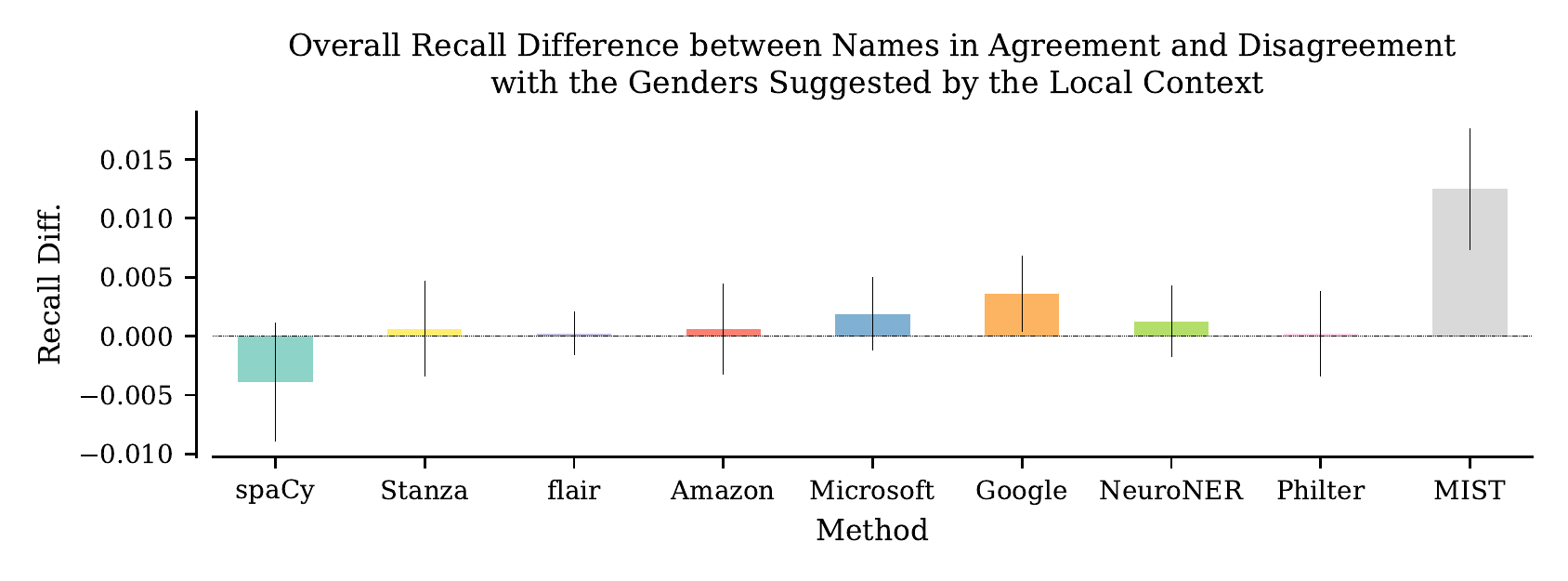}
    \vspace{-17.5pt}
    \caption{Difference in recall and 95\% bootstrapped confidence interval between names that are consistent and inconsistent with the genders suggested by the local context. A positive recall difference means that performance was best when there was gender consistency, while a negative recall difference means that performance was best when there was gender inconsistency. Methods leveraging the gender context for name recognition are expected to see a positive recall difference.}
    \label{fig:context_difference}
    \vspace{-7.5pt}
\end{figure*}

\begin{figure*}[t!]
    \centering
    \includegraphics[width=\textwidth]{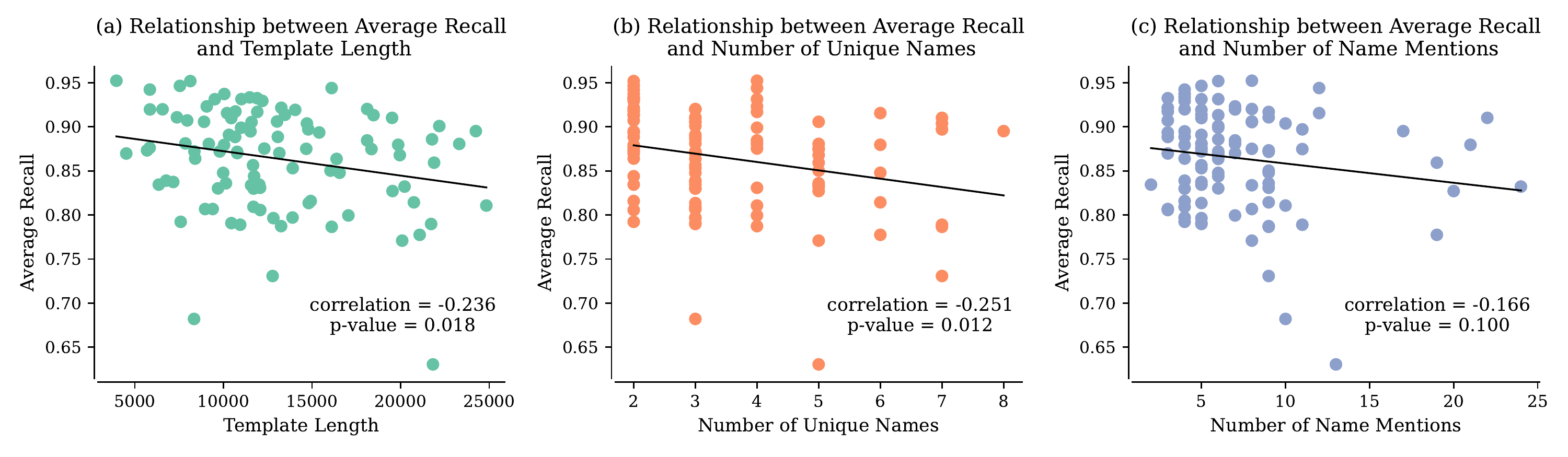}
    \vspace{-22.5pt}
    \caption{Relationship between template characteristics and template recall averaged across the examined methods. With statistically significant p-values, a template's average recall decreases with its length and the number of unique names included.}
    \label{fig:relationship_template}
    \vspace{-7.5pt}
\end{figure*}

\begin{figure*}[t!]
    \centering
    \includegraphics[width=\textwidth]{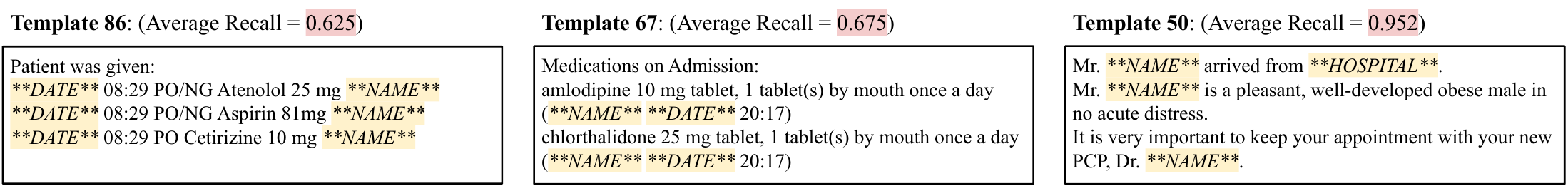}
    \vspace{-17.5pt}
    \caption{Average recall and snippets of three templates. Unlike usual templates (e.g., Template 50), templates with a low average recall (e.g., Templates 86 and 67) usually include medication injection histories that offer little semantic context for name recognition.}
    \label{fig:template_snippet}
    \vspace{-10pt}
\end{figure*}

As shown in Figure~\ref{fig:polysemy_recall}, although we utilize polysemy first names for all three racial groups, the variation in performance persists.
In addition, for all the methods except \texttt{Stanza}, the recall ranking of the three racial groups assessed on polysemy first names remains relatively consistent as that based on the original setting in Figure~\ref{fig:dimension_recall} (b).
We also consider the scenario when a method can correctly recognize the non-polysemy last names and plot the increased recall above the original bar in lighter colors in Figure~\ref{fig:polysemy_recall}.
In this case, most methods can see a significant increase in recall, especially for \texttt{Google}, \texttt{NeuroNER}, and \texttt{Philter}.
Hence, names with overlapping meanings in English only explain the underperformance of the de-identification methods, but not necessarily their bias across demographic groups.

\begin{table*}[t!]
  \centering
  \setlength\tabcolsep{4pt}
  \resizebox{\textwidth}{!}{
  \begin{tabular}{c|cc|ccc|cccc} 
    \toprule
         \multirow{2}{*}{\textbf{Method}} & \multicolumn{2}{c|}{\textbf{Fine-tuning}} & \multicolumn{3}{c|}{\textbf{Overall Performance ($\uparrow$)}} & \multicolumn{4}{c}{\textbf{Bias along Dimensions ($\downarrow$)}} \\
         & \textbf{Context} & \textbf{Name} & \textbf{Precision} & \textbf{Recall} & \textbf{F1} & \textbf{Gender} & \textbf{Race} & \textbf{Popularity} & \textbf{Decade} \\
    \midrule 
    \midrule
        \multirow{5}{*}{\texttt{spaCy}} & \multicolumn{2}{c|}{out-of-the-box} & 0.916 & 0.623 & 0.741 & \textbf{0.003} & 0.027 & 0.025 & 0.005 \\
        & clinical & diverse & 0.990$\pm$0.007 & \textbf{0.950}$\pm$\textbf{0.006} & \textbf{0.969}$\pm$\textbf{0.002} & 0.012$\pm$0.004 & \textbf{0.024}$\pm$\textbf{0.005} & \textbf{0.005}$\pm$\textbf{0.002} & 0.006$\pm$0.001 \\
        & clinical & popular & \textbf{0.998}$\pm$\textbf{0.004} & 0.737$\pm$0.072 & 0.846$\pm$0.046 & 0.012$\pm$0.007 & 0.094$\pm$0.029 & 0.127$\pm$0.035 & \textbf{0.003}$\pm$\textbf{0.004} \\
        & general & diverse & 0.915$\pm$0.072 & 0.830$\pm$0.083 & 0.864$\pm$0.035 & 0.036$\pm$0.005 & 0.071$\pm$0.011 & 0.049$\pm$0.042 & 0.008$\pm$0.005 \\
        & general & popular & 0.873$\pm$0.110 & 0.492$\pm$0.069 & 0.629$\pm$0.083 & 0.010$\pm$0.003 & 0.059$\pm$0.032 & 0.326$\pm$0.060 & 0.007$\pm$0.003 \\
    \midrule
        \multirow{5}{*}{\texttt{NeuroNER}} & \multicolumn{2}{c|}{out-of-the-box} & 0.955 & 0.953 & 0.954 & 0.005 & 0.044 & 0.030 & \textbf{0.001} \\
        & clinical & diverse & 0.978$\pm$0.014 & \textbf{0.978}$\pm$\textbf{0.009} & \textbf{0.978}$\pm$\textbf{0.005} & 0.007$\pm$0.001 & \textbf{0.019}$\pm$\textbf{0.006} & \textbf{0.012}$\pm$\textbf{0.008} & 0.002$\pm$0.001 \\
        & clinical & popular & \textbf{0.989}$\pm$\textbf{0.003} & 0.865$\pm$0.021 & 0.923$\pm$0.013 & 0.008$\pm$0.004 & 0.065$\pm$0.007 & 0.118$\pm$0.010 & \textbf{0.001}$\pm$\textbf{0.001} \\
        & general & diverse & 0.958$\pm$0.022 & 0.943$\pm$0.029 & 0.950$\pm$0.010 & 0.016$\pm$0.007 & 0.041$\pm$0.010 & 0.031$\pm$0.014 & 0.007$\pm$0.006 \\
        & general & popular & 0.924$\pm$0.022 & 0.777$\pm$0.018 & 0.844$\pm$0.019 & \textbf{0.003}$\pm$\textbf{0.001} & 0.062$\pm$0.005 & 0.324$\pm$0.021 & 0.004$\pm$0.003 \\
    \bottomrule 
  \end{tabular}
  }
  \caption{Overall performance (higher is better) and bias along demographic dimensions (lower is better) of two de-identification methods fine-tuned with different setups. We measure the bias with recall equality difference, report the mean scores and standard errors based on five trials with different seeds, and bold the best score in each column for each method. For both methods, using clinical context and diverse names for fine-tuning improves the overall performance and reduces the demographic bias along most dimensions, especially race and popularity.}
  \label{tab:finetune_performance} 
  \vspace{-22.5pt}
\end{table*}

\subsection{Methods Improve when De-identifying Context-Consistent Names} \label{sec:5.2}
NER systems usually capture the contextual dependencies for tag decoding \cite{li2020survey}, and the semantic context often indicates the gender associated with a name.
For example, titles (e.g., Mr. and Mrs.) can appear before full names, and appositions (e.g., son and daughter) can describe relationships.
We expect methods leveraging such context for name recognition to have higher recall on names where there is local context agreement with the gender as compared to those with disagreement.
To assess this, we identify in our note templates where name gender can be easily inferred from the local context to determine if the consistency between the names and the inferred genders impacts de-identification quality.

Figure~\ref{fig:context_difference} plots the recall difference between context-consistent and -inconsistent names by the examined methods. 
Albeit with relatively large confidence intervals, We find that most methods perform better on names aligned with the implied gender.
\texttt{spaCy} is the only exception, perhaps shedding light on its lowest overall recall (see Table~\ref{tab:overall_performance}).

\textit{Limitations of Gender-Inconsistent Evaluation in Experiment Setup.}
We acknowledge that replacing gender-inconsistent pronouns in notes prior to evaluation would be an easier test for models. 
However, we note that not all clinical records will contain gender-confirming pronouns, especially for transgender and non-binary individuals \cite{lockhart2023name}, and argue that de-identification methods should be able to operate properly in these gender-inconsistent situations.
We also note that if we limit our analysis to only using male-originating notes with male name sets and female-originating notes with female name sets, our results still hold (see Appendix~\ref{app:1.1}).
We note that in this setting, we do not explicitly assess the gender gap since male- and female-originating notes do not overlap.

\subsection{Performance Decays with Template Length and Name Quantity} \label{sec:5.3}
Other properties of a note template may also affect the de-identifica-tion performance.
We consider three characteristics---template leng-th, number of unique names, and number of name mentions in a template---and visualize their relationships with a template's average recall in Figure~\ref{fig:relationship_template}.
Our findings suggest that recall deteriorates with both the length of a note and the number of unique names that it contains.

We identify two of the worst-performing templates in terms of recall: Templates 86 and 67. 
These templates appear six and four times, respectively, in the five templates with the lowest recall by a method. 
As shown in Figure~\ref{fig:template_snippet}, unlike other templates (e.g., Template 50), Templates 86 and 67 are notable for having large blocks of medication history that provide little indication for the names that intersperse them. 
This unique characteristic of clinical records calls for special attention in future de-identification systems.
We further investigate the performance of the examined methods on these hard templates in Appendix~\ref{app:1.2} and find that their performance follows the overall pattern in Table~\ref{tab:overall_performance}.
\vspace{-4pt}
\section{Q3: Can Bias Be Mitigated?} \label{sec:6}
To answer the third question of how to mitigate the bias in de-identification methods, we propose a simple and method-agnostic solution of fine-tuning the methods with clinical context and diverse names.
This setup not only improves the overall recall but also reduces the bias significantly along most demographic dimensions.

\vspace{-4pt}
\subsection{Fine-tuning De-identification Methods} \label{sec:6.1}
We prepare the fine-tuning de-identification datasets by considering two types of context and two types of names.
We treat the longitudinal clinical narratives in the 2014 i2b2 de-identification challenge \cite{stubbs2015annotating} as the clinical context and the Wikipedia articles in the DocRED dataset \cite{yao2019DocRED} as the general context.
We generate 160 diverse names by randomly sampling ten names from each of the 16 name sets in Table~\ref{tab:name_set} and 160 popular names based on the most popular names over the three chosen decades that do not appear in the 16 name sets.
For each type of context, we randomly sample \numprint{1000} templates for training and 100 for validation.
These templates are then populated with the names of each type (i.e., diverse names and popular names) separately.
In this way, we create four fine-tuning setups in total by pairing the two types of context with the two types of names.

To compare the effectiveness of these setups, we fine-tune two de-identification methods---\texttt{spaCy} \cite{honnibal2020spacy} and \texttt{NeuroNER} \cite{dernoncourt2017neuroner, dernoncourt2016deidentification}---with distinct out-of-the-box performance.
\texttt{spaCy} is a widely-adopted NLP library that delivers a low de-identification recall and a moderate demographic bias in Table~\ref{tab:overall_performance}.
In contrast, \texttt{NeuroNER} is pre-trained on the original 2014 i2b2 de-identification corpus, which yields a competitive recall with high bias along the dimensions of race and popularity.
After fine-tuning with their respective default hyperparameters, these methods are evaluated on \numprint{1600} test notes.
These test notes are constructed by filling in the 100 templates in Sec~\ref{sec:3.4} with the remaining ten names (not selected for the 160 diverse names during fine-tuning) from each of the 16 name sets separately.
Here, the test notes are disjoint with the fine-tuning context/names.

\vspace{-4pt}
\subsection{Clinical Context and Diverse Names Improve Performance} \label{sec:6.2}
Table~\ref{tab:finetune_performance} displays the overall performance and the demographic bias (i.e., the recall equality difference) of the two methods after fine-tuning.
We repeat the fine-tuning five times with different seeds and report the mean scores and standard errors.
Impressively, despite the distinct out-of-the-box performance of the two fine-tuned methods, the setup composed of clinical context and diverse names largely enhances the overall performance of both methods and diminishes their unfairness, especially along the dimensions of race and popularity.

In particular, although most of the four fine-tuning setups improve \texttt{spaCy}'s overall performance, fine-tuning with clinical context and diverse names sees the largest boost in \texttt{spaCy}'s recall by over 0.3.
On the other hand, since \texttt{NeuroNER} is pre-trained on clinical corpora, most of the four fine-tuning setups are ineffective in enhancing \texttt{NeuroNER}'s strong out-of-the-box performance.
However, fine-tuning with clinical context and diverse names is the only exception here, which increases the precision, recall, and F1 of \texttt{NeuroNER} by around 0.02 each.
Moreover, along the dimensions of race and popularity, where the degree of unfairness is rather high, this setup can significantly reduce the bias of both methods.

We suggest that fine-tuning de-identification methods with clinical context and diverse names should be done as an immediate fix to improve fairness before applying the methods to clinical tasks. 
The method-agnostic effectiveness and simplicity of this setup highlight the importance of training data diversity to model fairness \cite{Madan2022when}.
\vspace{-3pt}
\section{Discussion} \label{sec:7}
\paragraph{Demographic Associations of Names}
Names can be associated with certain demographic features \cite{liu2013s, gaddis2017black}. 
For instance, in our U.S. Social Security \cite{ssaPopularBaby} and Census \cite{censusDecennialCensus} data sources, there is variation in name popularity between self-reported ethnic groups.
In human decision-making, such associations have been shown to correlate with discriminative hiring \cite{bertrand2004emily, hannak2017bias} and loan granting \cite{hanson2016discrimination} practices.
Other work has explored the biases learned by large language models when the demographic context is varied directly in input \cite{liang2021towards} or using names as a proxy for demographic \cite{mansfield2022behind, mehrabi2020man, mishra2020assessing}.
For example, NLP models link the female gender to specific stereotypical occupations \cite{bolukbasi2016man} and tend to generate violent or negative-toned text when given ``Muslim'' as a demographic descriptor for input \cite{abid2021persistent}.
We emphasize that the biases inherently learned by NLP models may perpetuate biases and, therefore, require careful audits. 
We acknowledge that our analysis based on de-identifying names may not necessarily generalize to other PHI types and leave this further investigation to future work.

\vspace{-2pt}
\paragraph{Bias in Healthcare}
Bias in healthcare can occur in both systematic and implicit ways based on demographic factors such as race, ethnicity, gender, sexual orientation, or socio-economic status \cite{fitzgerald2017implicit, zestcott2016examining, marcelin2019impact, hall2015implicit}. 
These biases can then be unintentionally learned by ML models \cite{ahmad2020fairness, gianfrancesco2018potential, parikh2019addressing}.
For instance, NLP models trained on race-redacted clinical notes have been shown to capture self-reported race through other proxy information \cite{adam2022write} and mimic the existing biases in text completions for clinical treatment decisions \cite{zhang2020hurtful}. 
Our study demonstrates that existing clinical de-identification methods discriminate based on the demographic associations of names.
The bias in these methods could further escalate the unfairness in downstream healthcare systems.

\vspace{-2pt}
\paragraph{Importance of De-identified Data for Reproducibility}
ML models rely on large amounts of data for training \cite{beam2018big}, but in the case of health data, there are privacy concerns.
By removing PHI, researchers can protect stakeholders' privacy with de-identified data \cite{seastedt2022global} and avoid biasing their models through more representative datasets \cite{chen2021ethical}. 
To this end, clinical de-identification has attracted long-lasting attention from the research community \cite{kayaalp2017modes, meystre2010automatic} and large amounts of resources from the industrial world (e.g, \cite{privacyanalyticsPrivacyAnalytics, healthverityHealthVerityCensus}).
We highlight the importance of equitable de-identification because legal and ethical data sharing should be encouraged \cite{seastedt2022global} to improve the reproducibility of clinical findings and the credibility of healthcare systems \cite{mcdermott2021reproducibility, tsima2023reproducibility}.

\vspace{-2pt}
\paragraph{Harm of Minority Exclusion}
We stress that it is not acceptable to exclude some populations from de-identified data sharing.
When demographic groups are absent in data, models trained on that data will perform poorly on the missing groups \cite{norori2021addressing}.
This can result in misdiagnoses, inadequate treatments, and a failure to address health disparities \cite{ghassemi2022medicine}.
Hence, it is crucial to ensure that data for model training is diverse and representative of the populations they will serve \cite{chen2021ethical}.
Future work should consider proactive measures to collect and include data from underrepresented populations and address systemic biases during data collection and analysis.

\vspace{-2pt}
\paragraph{Ramifications of Poorer Privacy for Marginalized Groups} 
General disparities in de-identification performance can lead to poorer privacy for marginalized groups and engender crimes such as identity theft \cite{anderson2005identity, anderson2006victims}.
This adds to the existing difficulties with data collection and monitoring faced by marginalized communities \cite{browne2015dark, fagan2016stops}.
Even when data sharing is consented, the data can be used outside of the given context, leading to representational harm for groups that are already targeted \cite{ghassemi2022machine}.
In future work, we advocate for data collection and de-identification practices that promote trust and do not discourage minorities from seeking medical care and participating in clinical data sharing. 

\vspace{-2pt}
\paragraph{Importance of Audits to Create Change}
Audits in healthcare help to identify areas of improvement \cite{ivers2012audit}, assess compliance with regulations and standards \cite{hut2021understanding}, and hold organizations accountable for their actions \cite{raji2019actionable}.
Past work on ML audits has demonstrated the ability to make meaningful changes and reduce performance gaps in deployed systems with biases.
For example, a recent audit on the bias in automated facial analysis algorithms \cite{buolamwini2018gender} stimulated the targeted companies to reduce accuracy disparities between demographic groups \cite{raji2019actionable}, 
However, companies that provided similar algorithms and were not included in the original audit did not make corresponding changes \cite{raji2019actionable}.
We encourage clinical practitioners to build upon our de-identification audit to provide high-quality, equitable de-identification services to all demographic groups.
\vspace{-2pt}
\section{Conclusion} \label{sec:8}
In this paper, we contribute a large-scale empirical analysis of de-identifying names from clinical records and present findings that demonstrate systemic bias in performance. 
Our results should sound the alarm for clinical and ML stakeholders, as bias in clinical de-identification not only raises legal concerns but also make certain demographic groups more prone to privacy leakage.
Hence, we call for an urgent review of existing de-identification methods and actions (e.g., fine-tuning with our recommended setup) to improve the fairness and accountability of healthcare systems.

Despite the comprehensiveness of our study, we acknowledge the limitation of using coarse racial and gender categorizations when constructing our name sets.
In addition, while our analysis is readily applicable to many widely-adopted de-identification methods, we did not evaluate its generalization to approaches focusing on other PHI classes.
We leave to future work the investigation of bias in de-identifying other PHI classes based on more fluid racial and gender categorizations.

\vspace{-2pt}
\begin{acks}
This project is supported by the National Institute of Biomedical Imaging and Bioengineering (NIBIB) under NIH grant number R01EB030362.
We would like to acknowledge the contributions of Dana Moukheiber, Lama Moukheiber, and Mira Moukheiber in annotating the clinical note templates used in our experiments.
\end{acks}

\newpage
\bibliographystyle{ACM-Reference-Format}
\bibliography{main}


\begin{thebibliography}{142}


\ifx \showCODEN    \undefined \def \showCODEN     #1{\unskip}     \fi
\ifx \showDOI      \undefined \def \showDOI       #1{#1}\fi
\ifx \showISBNx    \undefined \def \showISBNx     #1{\unskip}     \fi
\ifx \showISBNxiii \undefined \def \showISBNxiii  #1{\unskip}     \fi
\ifx \showISSN     \undefined \def \showISSN      #1{\unskip}     \fi
\ifx \showLCCN     \undefined \def \showLCCN      #1{\unskip}     \fi
\ifx \shownote     \undefined \def \shownote      #1{#1}          \fi
\ifx \showarticletitle \undefined \def \showarticletitle #1{#1}   \fi
\ifx \showURL      \undefined \def \showURL       {\relax}        \fi
\providecommand\bibfield[2]{#2}
\providecommand\bibinfo[2]{#2}
\providecommand\natexlab[1]{#1}
\providecommand\showeprint[2][]{arXiv:#2}

\bibitem[goo(nd)]%
        {googleDeidentifyingSensitive}
 \bibinfo{year}{[n.d.]}\natexlab{}.
\newblock \bibinfo{title}{{D}e-identifying sensitive data | {D}ata {L}oss {P}revention {D}ocumentation | {G}oogle {C}loud --- cloud.google.com}.
\newblock \bibinfo{howpublished}{\url{https://cloud.google.com/dlp/docs/deidentify-sensitive-data}}.
\newblock
\newblock
\shownote{[Accessed 24-November-2022]}.


\bibitem[ggH(nd)]%
        {ggHealthcareDeidentify}
 \bibinfo{year}{[n.d.]}\natexlab{}.
\newblock \bibinfo{title}{De-identifying sensitive data  |  cloud healthcare API  |  google cloud}.
\newblock \bibinfo{howpublished}{\url{https://cloud.google.com/healthcare-api/docs/how-tos/deidentify}}.
\newblock
\newblock
\shownote{[Accessed 24-November-2022]}.


\bibitem[cen(nd)]%
        {censusDecennialCensus}
 \bibinfo{year}{[n.d.]}\natexlab{}.
\newblock \bibinfo{title}{{D}ecennial {C}ensus {S}urname {F}iles (2010, 2000) --- census.gov}.
\newblock \bibinfo{howpublished}{\url{https://www.census.gov/data/developers/data-sets/surnames.html}}.
\newblock
\newblock
\shownote{[Accessed 30-June-2022]}.


\bibitem[ama(nd)]%
        {amazonDetectAmazon}
 \bibinfo{year}{[n.d.]}\natexlab{}.
\newblock \bibinfo{title}{{D}etect {P}{H}{I} - {A}mazon {C}omprehend {M}edical --- docs.aws.amazon.com}.
\newblock \bibinfo{howpublished}{\url{https://docs.aws.amazon.com/comprehend-medical/latest/dev/textanalysis-phi.html}}.
\newblock
\newblock
\shownote{[Accessed 24-November-2022]}.


\bibitem[hea(nd)]%
        {healthverityHealthVerityCensus}
 \bibinfo{year}{[n.d.]}\natexlab{}.
\newblock \bibinfo{title}{{H}ealth{V}erity {C}ensus – {R}eal-{T}ime {P}atient {I}dentity {R}esolution {T}echnology --- healthverity.com}.
\newblock \bibinfo{howpublished}{\url{https://healthverity.com/solutions/healthverity-census/}}.
\newblock
\newblock
\shownote{[Accessed 06-Feb-2023]}.


\bibitem[azu(nd)]%
        {azureGPT}
 \bibinfo{year}{[n.d.]}\natexlab{}.
\newblock \bibinfo{title}{{H}ow to work with the {G}{P}{T}-35-{T}urbo and {G}{P}{T}-4 models - {A}zure {O}pen{A}{I} {S}ervice --- learn.microsoft.com}.
\newblock \bibinfo{howpublished}{\url{https://learn.microsoft.com/en-us/azure/ai-services/openai/how-to/chatgpt?tabs=python-new&pivots=programming-language-chat-completions}}.
\newblock
\newblock
\shownote{[Accessed 29-November-2023]}.


\bibitem[ssa(nd)]%
        {ssaPopularBaby}
 \bibinfo{year}{[n.d.]}\natexlab{}.
\newblock \bibinfo{title}{{P}opular {B}aby {N}ames --- ssa.gov}.
\newblock \bibinfo{howpublished}{\url{https://www.ssa.gov/oact/babynames/limits.html}}.
\newblock
\newblock
\shownote{[Accessed 30-June-2022]}.


\bibitem[pri(nd)]%
        {privacyanalyticsPrivacyAnalytics}
 \bibinfo{year}{[n.d.]}\natexlab{}.
\newblock \bibinfo{title}{{P}rivacy {A}nalytics - {S}oftware to {A}nonymize {T}ext --- privacy-analytics.com}.
\newblock \bibinfo{howpublished}{\url{https://privacy-analytics.com/health-data-privacy/health-data-software/software-to-anonymize-text/}}.
\newblock
\newblock
\shownote{[Accessed 06-Feb-2023]}.


\bibitem[gHe(nd)]%
        {gHealthcareNLP}
 \bibinfo{year}{[n.d.]}\natexlab{}.
\newblock \bibinfo{title}{Using the healthcare natural language API | cloud healthcare API | google cloud}.
\newblock \bibinfo{howpublished}{\url{https://cloud.google.com/healthcare-api/docs/how-tos/nlp}}.
\newblock
\newblock
\shownote{[Accessed 24-November-2022]}.


\bibitem[mic(nd)]%
        {microsoftWhatPersonally}
 \bibinfo{year}{[n.d.]}\natexlab{}.
\newblock \bibinfo{title}{{W}hat is the {P}ersonally {I}dentifying {I}nformation ({P}{I}{I}) detection feature in {A}zure {C}ognitive {S}ervice for {L}anguage? - {A}zure {C}ognitive {S}ervices --- learn.microsoft.com}.
\newblock \bibinfo{howpublished}{\url{https://learn.microsoft.com/en-us/azure/cognitive-services/language-service/personally-identifiable-information/overview}}.
\newblock
\newblock
\shownote{[Accessed 24-November-2022]}.


\bibitem[Aberdeen et~al\mbox{.}(2010)]%
        {aberdeen2010mitre}
\bibfield{author}{\bibinfo{person}{John Aberdeen}, \bibinfo{person}{Samuel Bayer}, \bibinfo{person}{Reyyan Yeniterzi}, \bibinfo{person}{Ben Wellner}, \bibinfo{person}{Cheryl Clark}, \bibinfo{person}{David Hanauer}, \bibinfo{person}{Bradley Malin}, {and} \bibinfo{person}{Lynette Hirschman}.} \bibinfo{year}{2010}\natexlab{}.
\newblock \showarticletitle{The MITRE Identification Scrubber Toolkit: design, training, and assessment}.
\newblock \bibinfo{journal}{\emph{International journal of medical informatics}} (\bibinfo{year}{2010}).
\newblock


\bibitem[Abid et~al\mbox{.}(2021)]%
        {abid2021persistent}
\bibfield{author}{\bibinfo{person}{Abubakar Abid}, \bibinfo{person}{Maheen Farooqi}, {and} \bibinfo{person}{James Zou}.} \bibinfo{year}{2021}\natexlab{}.
\newblock \showarticletitle{Persistent anti-muslim bias in large language models}. In \bibinfo{booktitle}{\emph{Proceedings of the 2021 AAAI/ACM Conference on AI, Ethics, and Society}}.
\newblock


\bibitem[Adam et~al\mbox{.}(2022)]%
        {adam2022write}
\bibfield{author}{\bibinfo{person}{Hammaad Adam}, \bibinfo{person}{Ming~Ying Yang}, \bibinfo{person}{Kenrick Cato}, \bibinfo{person}{Ioana Baldini}, \bibinfo{person}{Charles Senteio}, \bibinfo{person}{Leo~Anthony Celi}, \bibinfo{person}{Jiaming Zeng}, \bibinfo{person}{Moninder Singh}, {and} \bibinfo{person}{Marzyeh Ghassemi}.} \bibinfo{year}{2022}\natexlab{}.
\newblock \showarticletitle{Write It Like You See It: Detectable Differences in Clinical Notes by Race Lead to Differential Model Recommendations}. In \bibinfo{booktitle}{\emph{Proceedings of the 2022 AAAI/ACM Conference on AI, Ethics, and Society}}.
\newblock


\bibitem[Agrawal et~al\mbox{.}(2022)]%
        {agrawal2022large}
\bibfield{author}{\bibinfo{person}{Monica Agrawal}, \bibinfo{person}{Stefan Hegselmann}, \bibinfo{person}{Hunter Lang}, \bibinfo{person}{Yoon Kim}, {and} \bibinfo{person}{David Sontag}.} \bibinfo{year}{2022}\natexlab{}.
\newblock \showarticletitle{Large language models are few-shot clinical information extractors}. In \bibinfo{booktitle}{\emph{Proceedings of the 2022 Conference on Empirical Methods in Natural Language Processing}}. \bibinfo{pages}{1998--2022}.
\newblock


\bibitem[Ahmad et~al\mbox{.}(2018)]%
        {ahmad2018interpretable}
\bibfield{author}{\bibinfo{person}{Muhammad~Aurangzeb Ahmad}, \bibinfo{person}{Carly Eckert}, {and} \bibinfo{person}{Ankur Teredesai}.} \bibinfo{year}{2018}\natexlab{}.
\newblock \showarticletitle{Interpretable machine learning in healthcare}. In \bibinfo{booktitle}{\emph{Proceedings of the 2018 ACM international conference on bioinformatics, computational biology, and health informatics}}.
\newblock


\bibitem[Ahmad et~al\mbox{.}(2020)]%
        {ahmad2020fairness}
\bibfield{author}{\bibinfo{person}{Muhammad~Aurangzeb Ahmad}, \bibinfo{person}{Arpit Patel}, \bibinfo{person}{Carly Eckert}, \bibinfo{person}{Vikas Kumar}, {and} \bibinfo{person}{Ankur Teredesai}.} \bibinfo{year}{2020}\natexlab{}.
\newblock \showarticletitle{Fairness in machine learning for healthcare}. In \bibinfo{booktitle}{\emph{Proceedings of the 26th ACM SIGKDD International Conference on Knowledge Discovery \& Data Mining}}.
\newblock


\bibitem[Akbik et~al\mbox{.}(2019)]%
        {akbik2019flair}
\bibfield{author}{\bibinfo{person}{Alan Akbik}, \bibinfo{person}{Tanja Bergmann}, \bibinfo{person}{Duncan Blythe}, \bibinfo{person}{Kashif Rasul}, \bibinfo{person}{Stefan Schweter}, {and} \bibinfo{person}{Roland Vollgraf}.} \bibinfo{year}{2019}\natexlab{}.
\newblock \showarticletitle{FLAIR: An easy-to-use framework for state-of-the-art NLP}. In \bibinfo{booktitle}{\emph{Proceedings of the 2019 conference of the North American chapter of the association for computational linguistics (demonstrations)}}.
\newblock


\bibitem[Akbik et~al\mbox{.}(2018)]%
        {akbik2018contextual}
\bibfield{author}{\bibinfo{person}{Alan Akbik}, \bibinfo{person}{Duncan Blythe}, {and} \bibinfo{person}{Roland Vollgraf}.} \bibinfo{year}{2018}\natexlab{}.
\newblock \showarticletitle{Contextual string embeddings for sequence labeling}. In \bibinfo{booktitle}{\emph{Proceedings of the 27th international conference on computational linguistics}}.
\newblock


\bibitem[Anderson(2005)]%
        {anderson2005identity}
\bibfield{author}{\bibinfo{person}{Keith~B Anderson}.} \bibinfo{year}{2005}\natexlab{}.
\newblock \showarticletitle{Identity theft: Does the risk vary with demographics?}
\newblock \bibinfo{journal}{\emph{Federal Trade Commission, Bureau of Economics Working Paper}} (\bibinfo{year}{2005}).
\newblock


\bibitem[Anderson(2006)]%
        {anderson2006victims}
\bibfield{author}{\bibinfo{person}{Keith~B Anderson}.} \bibinfo{year}{2006}\natexlab{}.
\newblock \showarticletitle{Who are the victims of identity theft? The effect of demographics}.
\newblock \bibinfo{journal}{\emph{Journal of Public Policy \& Marketing}} (\bibinfo{year}{2006}).
\newblock


\bibitem[Beam and Kohane(2018)]%
        {beam2018big}
\bibfield{author}{\bibinfo{person}{Andrew~L Beam} {and} \bibinfo{person}{Isaac~S Kohane}.} \bibinfo{year}{2018}\natexlab{}.
\newblock \showarticletitle{Big data and machine learning in health care}.
\newblock \bibinfo{journal}{\emph{Jama}} (\bibinfo{year}{2018}).
\newblock


\bibitem[Beckwith et~al\mbox{.}(2006)]%
        {beckwith2006development}
\bibfield{author}{\bibinfo{person}{Bruce~A Beckwith}, \bibinfo{person}{Rajeshwarri Mahaadevan}, \bibinfo{person}{Ulysses~J Balis}, {and} \bibinfo{person}{Frank Kuo}.} \bibinfo{year}{2006}\natexlab{}.
\newblock \showarticletitle{Development and evaluation of an open source software tool for deidentification of pathology reports}.
\newblock \bibinfo{journal}{\emph{BMC medical informatics and decision making}} (\bibinfo{year}{2006}).
\newblock


\bibitem[Bergdall et~al\mbox{.}(2012)]%
        {bergdall2012cb3}
\bibfield{author}{\bibinfo{person}{Anna Bergdall}, \bibinfo{person}{Stephen Asche}, \bibinfo{person}{Nicole Schneider}, \bibinfo{person}{Tessa Kerby}, \bibinfo{person}{Karen Margolis}, \bibinfo{person}{JoAnn Sperl-Hillen}, \bibinfo{person}{Jaime Sekenski}, \bibinfo{person}{Rachel Pritchard}, \bibinfo{person}{Michael Maciosek}, {and} \bibinfo{person}{Patrick O’Connor}.} \bibinfo{year}{2012}\natexlab{}.
\newblock \showarticletitle{CB3-01: comparison of ethnicity and race categorization in electronic medical records and by Self-report}.
\newblock \bibinfo{journal}{\emph{Clinical Medicine \& Research}} (\bibinfo{year}{2012}).
\newblock


\bibitem[Bertrand and Mullainathan(2004)]%
        {bertrand2004emily}
\bibfield{author}{\bibinfo{person}{Marianne Bertrand} {and} \bibinfo{person}{Sendhil Mullainathan}.} \bibinfo{year}{2004}\natexlab{}.
\newblock \showarticletitle{Are Emily and Greg more employable than Lakisha and Jamal? A field experiment on labor market discrimination}.
\newblock \bibinfo{journal}{\emph{American economic review}} (\bibinfo{year}{2004}).
\newblock


\bibitem[Beutel et~al\mbox{.}(2019)]%
        {beutel2019putting}
\bibfield{author}{\bibinfo{person}{Alex Beutel}, \bibinfo{person}{Jilin Chen}, \bibinfo{person}{Tulsee Doshi}, \bibinfo{person}{Hai Qian}, \bibinfo{person}{Allison Woodruff}, \bibinfo{person}{Christine Luu}, \bibinfo{person}{Pierre Kreitmann}, \bibinfo{person}{Jonathan Bischof}, {and} \bibinfo{person}{Ed~H Chi}.} \bibinfo{year}{2019}\natexlab{}.
\newblock \showarticletitle{Putting fairness principles into practice: Challenges, metrics, and improvements}. In \bibinfo{booktitle}{\emph{Proceedings of the 2019 AAAI/ACM Conference on AI, Ethics, and Society}}.
\newblock


\bibitem[Bhaskaran and Bhallamudi(2019)]%
        {bhaskaran2019good}
\bibfield{author}{\bibinfo{person}{Jayadev Bhaskaran} {and} \bibinfo{person}{Isha Bhallamudi}.} \bibinfo{year}{2019}\natexlab{}.
\newblock \showarticletitle{Good Secretaries, Bad Truck Drivers? Occupational Gender Stereotypes in Sentiment Analysis}. In \bibinfo{booktitle}{\emph{Proceedings of the First Workshop on Gender Bias in Natural Language Processing}}.
\newblock


\bibitem[Bliss(2012)]%
        {bliss2012race}
\bibfield{author}{\bibinfo{person}{Catherine Bliss}.} \bibinfo{year}{2012}\natexlab{}.
\newblock \bibinfo{booktitle}{\emph{Race decoded: The genomic fight for social justice}}.
\newblock \bibinfo{publisher}{Stanford University Press}.
\newblock


\bibitem[Blodgett et~al\mbox{.}(2020)]%
        {blodgett2020language}
\bibfield{author}{\bibinfo{person}{Su~Lin Blodgett}, \bibinfo{person}{Solon Barocas}, \bibinfo{person}{Hal Daum{\'e}~III}, {and} \bibinfo{person}{Hanna Wallach}.} \bibinfo{year}{2020}\natexlab{}.
\newblock \showarticletitle{Language (Technology) is Power: A Critical Survey of “Bias” in NLP}. In \bibinfo{booktitle}{\emph{Proceedings of the 58th Annual Meeting of the Association for Computational Linguistics}}.
\newblock


\bibitem[Blodgett and O'Connor(2017)]%
        {blodgett2017racial}
\bibfield{author}{\bibinfo{person}{Su~Lin Blodgett} {and} \bibinfo{person}{Brendan O'Connor}.} \bibinfo{year}{2017}\natexlab{}.
\newblock \showarticletitle{Racial disparity in natural language processing: A case study of social media african-american english}.
\newblock \bibinfo{journal}{\emph{arXiv preprint arXiv:1707.00061}} (\bibinfo{year}{2017}).
\newblock


\bibitem[Bolukbasi et~al\mbox{.}(2016)]%
        {bolukbasi2016man}
\bibfield{author}{\bibinfo{person}{Tolga Bolukbasi}, \bibinfo{person}{Kai-Wei Chang}, \bibinfo{person}{James~Y Zou}, \bibinfo{person}{Venkatesh Saligrama}, {and} \bibinfo{person}{Adam~T Kalai}.} \bibinfo{year}{2016}\natexlab{}.
\newblock \showarticletitle{Man is to computer programmer as woman is to homemaker? debiasing word embeddings}.
\newblock \bibinfo{journal}{\emph{Advances in neural information processing systems}} (\bibinfo{year}{2016}).
\newblock


\bibitem[Borkan et~al\mbox{.}(2019)]%
        {borkan2019nuanced}
\bibfield{author}{\bibinfo{person}{Daniel Borkan}, \bibinfo{person}{Lucas Dixon}, \bibinfo{person}{Jeffrey Sorensen}, \bibinfo{person}{Nithum Thain}, {and} \bibinfo{person}{Lucy Vasserman}.} \bibinfo{year}{2019}\natexlab{}.
\newblock \showarticletitle{Nuanced metrics for measuring unintended bias with real data for text classification}. In \bibinfo{booktitle}{\emph{Companion proceedings of the 2019 world wide web conference}}.
\newblock


\bibitem[Browne(2015)]%
        {browne2015dark}
\bibfield{author}{\bibinfo{person}{Simone Browne}.} \bibinfo{year}{2015}\natexlab{}.
\newblock \bibinfo{booktitle}{\emph{Dark matters: On the surveillance of blackness}}.
\newblock \bibinfo{publisher}{Duke University Press}.
\newblock


\bibitem[Buolamwini and Gebru(2018)]%
        {buolamwini2018gender}
\bibfield{author}{\bibinfo{person}{Joy Buolamwini} {and} \bibinfo{person}{Timnit Gebru}.} \bibinfo{year}{2018}\natexlab{}.
\newblock \showarticletitle{Gender shades: Intersectional accuracy disparities in commercial gender classification}. In \bibinfo{booktitle}{\emph{Conference on fairness, accountability and transparency}}.
\newblock


\bibitem[Byrne and Tanesini(2015)]%
        {byrne2015instilling}
\bibfield{author}{\bibinfo{person}{Aidan Byrne} {and} \bibinfo{person}{Alessandra Tanesini}.} \bibinfo{year}{2015}\natexlab{}.
\newblock \showarticletitle{Instilling new habits: addressing implicit bias in healthcare professionals}.
\newblock \bibinfo{journal}{\emph{Advances in Health Sciences Education}} (\bibinfo{year}{2015}).
\newblock


\bibitem[Caliskan et~al\mbox{.}(2017)]%
        {caliskan2017semantics}
\bibfield{author}{\bibinfo{person}{Aylin Caliskan}, \bibinfo{person}{Joanna~J Bryson}, {and} \bibinfo{person}{Arvind Narayanan}.} \bibinfo{year}{2017}\natexlab{}.
\newblock \showarticletitle{Semantics derived automatically from language corpora contain human-like biases}.
\newblock \bibinfo{journal}{\emph{Science}} (\bibinfo{year}{2017}).
\newblock


\bibitem[Cao et~al\mbox{.}(2022)]%
        {cao2022generalizability}
\bibfield{author}{\bibinfo{person}{Jie Cao}, \bibinfo{person}{Xiaosong Zhang}, \bibinfo{person}{Vahakn Shahinian}, \bibinfo{person}{Huiying Yin}, \bibinfo{person}{Diane Steffick}, \bibinfo{person}{Rajiv Saran}, \bibinfo{person}{Susan Crowley}, \bibinfo{person}{Michael Mathis}, \bibinfo{person}{Girish~N Nadkarni}, \bibinfo{person}{Michael Heung}, {et~al\mbox{.}}} \bibinfo{year}{2022}\natexlab{}.
\newblock \showarticletitle{Generalizability of an acute kidney injury prediction model across health systems}.
\newblock \bibinfo{journal}{\emph{Nature Machine Intelligence}} (\bibinfo{year}{2022}).
\newblock


\bibitem[Cassidy et~al\mbox{.}(1999)]%
        {cassidy1999inferring}
\bibfield{author}{\bibinfo{person}{Kimberly~Wright Cassidy}, \bibinfo{person}{Michael~H Kelly}, {and} \bibinfo{person}{Lee'at~J Sharoni}.} \bibinfo{year}{1999}\natexlab{}.
\newblock \showarticletitle{Inferring gender from name phonology.}
\newblock \bibinfo{journal}{\emph{Journal of Experimental Psychology: General}} (\bibinfo{year}{1999}).
\newblock


\bibitem[Chaloner and Maldonado(2019)]%
        {chaloner2019measuring}
\bibfield{author}{\bibinfo{person}{Kaytlin Chaloner} {and} \bibinfo{person}{Alfredo Maldonado}.} \bibinfo{year}{2019}\natexlab{}.
\newblock \showarticletitle{Measuring gender bias in word embeddings across domains and discovering new gender bias word categories}. In \bibinfo{booktitle}{\emph{Proceedings of the First Workshop on Gender Bias in Natural Language Processing}}.
\newblock


\bibitem[Chen et~al\mbox{.}(2021)]%
        {chen2021ethical}
\bibfield{author}{\bibinfo{person}{Irene~Y Chen}, \bibinfo{person}{Emma Pierson}, \bibinfo{person}{Sherri Rose}, \bibinfo{person}{Shalmali Joshi}, \bibinfo{person}{Kadija Ferryman}, {and} \bibinfo{person}{Marzyeh Ghassemi}.} \bibinfo{year}{2021}\natexlab{}.
\newblock \showarticletitle{Ethical machine learning in healthcare}.
\newblock \bibinfo{journal}{\emph{Annual review of biomedical data science}} (\bibinfo{year}{2021}).
\newblock


\bibitem[Chouldechova and Roth(2020)]%
        {chouldechova2020snapshot}
\bibfield{author}{\bibinfo{person}{Alexandra Chouldechova} {and} \bibinfo{person}{Aaron Roth}.} \bibinfo{year}{2020}\natexlab{}.
\newblock \showarticletitle{A snapshot of the frontiers of fairness in machine learning}.
\newblock \bibinfo{journal}{\emph{Commun. ACM}} (\bibinfo{year}{2020}).
\newblock


\bibitem[Conneau et~al\mbox{.}(2020)]%
        {conneau2020unsupervised}
\bibfield{author}{\bibinfo{person}{Alexis Conneau}, \bibinfo{person}{Kartikay Khandelwal}, \bibinfo{person}{Naman Goyal}, \bibinfo{person}{Vishrav Chaudhary}, \bibinfo{person}{Guillaume Wenzek}, \bibinfo{person}{Francisco Guzm{\'a}n}, \bibinfo{person}{{\'E}douard Grave}, \bibinfo{person}{Myle Ott}, \bibinfo{person}{Luke Zettlemoyer}, {and} \bibinfo{person}{Veselin Stoyanov}.} \bibinfo{year}{2020}\natexlab{}.
\newblock \showarticletitle{Unsupervised Cross-lingual Representation Learning at Scale}. In \bibinfo{booktitle}{\emph{Proceedings of the 58th Annual Meeting of the Association for Computational Linguistics}}.
\newblock


\bibitem[Czarnowska et~al\mbox{.}(2021)]%
        {czarnowska2021quantifying}
\bibfield{author}{\bibinfo{person}{Paula Czarnowska}, \bibinfo{person}{Yogarshi Vyas}, {and} \bibinfo{person}{Kashif Shah}.} \bibinfo{year}{2021}\natexlab{}.
\newblock \showarticletitle{Quantifying social biases in nlp: A generalization and empirical comparison of extrinsic fairness metrics}.
\newblock \bibinfo{journal}{\emph{Transactions of the Association for Computational Linguistics}} (\bibinfo{year}{2021}).
\newblock


\bibitem[Davidson et~al\mbox{.}(2019)]%
        {davidson2019racial}
\bibfield{author}{\bibinfo{person}{Thomas Davidson}, \bibinfo{person}{Debasmita Bhattacharya}, {and} \bibinfo{person}{Ingmar Weber}.} \bibinfo{year}{2019}\natexlab{}.
\newblock \showarticletitle{Racial Bias in Hate Speech and Abusive Language Detection Datasets}. In \bibinfo{booktitle}{\emph{Proceedings of the Third Workshop on Abusive Language Online}}.
\newblock


\bibitem[De-Arteaga et~al\mbox{.}(2019)]%
        {de2019bias}
\bibfield{author}{\bibinfo{person}{Maria De-Arteaga}, \bibinfo{person}{Alexey Romanov}, \bibinfo{person}{Hanna Wallach}, \bibinfo{person}{Jennifer Chayes}, \bibinfo{person}{Christian Borgs}, \bibinfo{person}{Alexandra Chouldechova}, \bibinfo{person}{Sahin Geyik}, \bibinfo{person}{Krishnaram Kenthapadi}, {and} \bibinfo{person}{Adam~Tauman Kalai}.} \bibinfo{year}{2019}\natexlab{}.
\newblock \showarticletitle{Bias in bios: A case study of semantic representation bias in a high-stakes setting}. In \bibinfo{booktitle}{\emph{proceedings of the Conference on Fairness, Accountability, and Transparency}}. \bibinfo{pages}{120--128}.
\newblock


\bibitem[Dernoncourt et~al\mbox{.}(2017)]%
        {dernoncourt2017neuroner}
\bibfield{author}{\bibinfo{person}{Franck Dernoncourt}, \bibinfo{person}{Ji~Young Lee}, {and} \bibinfo{person}{Peter Szolovits}.} \bibinfo{year}{2017}\natexlab{}.
\newblock \showarticletitle{NeuroNER: an easy-to-use program for named-entity recognition based on neural networks}. In \bibinfo{booktitle}{\emph{Proceedings of the 2017 Conference on Empirical Methods in Natural Language Processing: System Demonstrations}}.
\newblock


\bibitem[Dernoncourt et~al\mbox{.}(2016)]%
        {dernoncourt2016deidentification}
\bibfield{author}{\bibinfo{person}{Franck Dernoncourt}, \bibinfo{person}{Ji~Young Lee}, \bibinfo{person}{Ozlem Uzuner}, {and} \bibinfo{person}{Peter Szolovits}.} \bibinfo{year}{2016}\natexlab{}.
\newblock \showarticletitle{De-identification of Patient Notes with Recurrent Neural Networks}.
\newblock \bibinfo{journal}{\emph{Journal of the American Medical Informatics Association (JAMIA)}} (\bibinfo{year}{2016}).
\newblock


\bibitem[Drozdowski et~al\mbox{.}(2020)]%
        {drozdowski2020demographic}
\bibfield{author}{\bibinfo{person}{Pawel Drozdowski}, \bibinfo{person}{Christian Rathgeb}, \bibinfo{person}{Antitza Dantcheva}, \bibinfo{person}{Naser Damer}, {and} \bibinfo{person}{Christoph Busch}.} \bibinfo{year}{2020}\natexlab{}.
\newblock \showarticletitle{Demographic bias in biometrics: A survey on an emerging challenge}.
\newblock \bibinfo{journal}{\emph{IEEE Transactions on Technology and Society}} (\bibinfo{year}{2020}).
\newblock


\bibitem[Eisenman(1995)]%
        {eisenman1995there}
\bibfield{author}{\bibinfo{person}{Russell Eisenman}.} \bibinfo{year}{1995}\natexlab{}.
\newblock \showarticletitle{Is there bias in US law enforcement?}
\newblock \bibinfo{journal}{\emph{The Journal of Social, Political, and Economic Studies}} (\bibinfo{year}{1995}).
\newblock


\bibitem[Fagan et~al\mbox{.}(2016)]%
        {fagan2016stops}
\bibfield{author}{\bibinfo{person}{Jeffrey Fagan}, \bibinfo{person}{Anthony~A Braga}, \bibinfo{person}{Rod~K Brunson}, {and} \bibinfo{person}{April Pattavina}.} \bibinfo{year}{2016}\natexlab{}.
\newblock \showarticletitle{Stops and stares: Street stops, surveillance, and race in the new policing}.
\newblock \bibinfo{journal}{\emph{Fordham Urb. LJ}} (\bibinfo{year}{2016}).
\newblock


\bibitem[FitzGerald and Hurst(2017)]%
        {fitzgerald2017implicit}
\bibfield{author}{\bibinfo{person}{Chlo{\"e} FitzGerald} {and} \bibinfo{person}{Samia Hurst}.} \bibinfo{year}{2017}\natexlab{}.
\newblock \showarticletitle{Implicit bias in healthcare professionals: a systematic review}.
\newblock \bibinfo{journal}{\emph{BMC medical ethics}} (\bibinfo{year}{2017}).
\newblock


\bibitem[Fleurence et~al\mbox{.}(2014)]%
        {fleurence2014launching}
\bibfield{author}{\bibinfo{person}{Rachael~L Fleurence}, \bibinfo{person}{Lesley~H Curtis}, \bibinfo{person}{Robert~M Califf}, \bibinfo{person}{Richard Platt}, \bibinfo{person}{Joe~V Selby}, {and} \bibinfo{person}{Jeffrey~S Brown}.} \bibinfo{year}{2014}\natexlab{}.
\newblock \showarticletitle{Launching PCORnet, a national patient-centered clinical research network}.
\newblock \bibinfo{journal}{\emph{Journal of the American Medical Informatics Association}} (\bibinfo{year}{2014}).
\newblock


\bibitem[Friedlin and McDonald(2008)]%
        {friedlin2008software}
\bibfield{author}{\bibinfo{person}{F~Jeff Friedlin} {and} \bibinfo{person}{Clement~J McDonald}.} \bibinfo{year}{2008}\natexlab{}.
\newblock \showarticletitle{A software tool for removing patient identifying information from clinical documents}.
\newblock \bibinfo{journal}{\emph{Journal of the American Medical Informatics Association}} (\bibinfo{year}{2008}).
\newblock


\bibitem[Friedman(1937)]%
        {friedman1937use}
\bibfield{author}{\bibinfo{person}{Milton Friedman}.} \bibinfo{year}{1937}\natexlab{}.
\newblock \showarticletitle{The use of ranks to avoid the assumption of normality implicit in the analysis of variance}.
\newblock \bibinfo{journal}{\emph{Journal of the american statistical association}} (\bibinfo{year}{1937}).
\newblock


\bibitem[Gaddis(2017)]%
        {gaddis2017black}
\bibfield{author}{\bibinfo{person}{S~Michael Gaddis}.} \bibinfo{year}{2017}\natexlab{}.
\newblock \showarticletitle{How black are Lakisha and Jamal? Racial perceptions from names used in correspondence audit studies}.
\newblock \bibinfo{journal}{\emph{Sociological Science}} (\bibinfo{year}{2017}).
\newblock


\bibitem[Ganz et~al\mbox{.}(2021)]%
        {ganz2021assessing}
\bibfield{author}{\bibinfo{person}{Melanie Ganz}, \bibinfo{person}{Sune~H Holm}, {and} \bibinfo{person}{Aasa Feragen}.} \bibinfo{year}{2021}\natexlab{}.
\newblock \showarticletitle{Assessing bias in medical ai}. In \bibinfo{booktitle}{\emph{Workshop on Interpretable ML in Healthcare at International Connference on Machine Learning (ICML)}}.
\newblock


\bibitem[Ghassemi and Mohamed(2022)]%
        {ghassemi2022machine}
\bibfield{author}{\bibinfo{person}{Marzyeh Ghassemi} {and} \bibinfo{person}{Shakir Mohamed}.} \bibinfo{year}{2022}\natexlab{}.
\newblock \showarticletitle{Machine learning and health need better values}.
\newblock \bibinfo{journal}{\emph{npj Digital Medicine}} (\bibinfo{year}{2022}).
\newblock


\bibitem[Ghassemi et~al\mbox{.}(2020)]%
        {ghassemi2020review}
\bibfield{author}{\bibinfo{person}{Marzyeh Ghassemi}, \bibinfo{person}{Tristan Naumann}, \bibinfo{person}{Peter Schulam}, \bibinfo{person}{Andrew~L Beam}, \bibinfo{person}{Irene~Y Chen}, {and} \bibinfo{person}{Rajesh Ranganath}.} \bibinfo{year}{2020}\natexlab{}.
\newblock \showarticletitle{A Review of Challenges and Opportunities in Machine Learning for Health}.
\newblock \bibinfo{journal}{\emph{AMIA Summits on Translational Science Proceedings}} (\bibinfo{year}{2020}).
\newblock


\bibitem[Ghassemi and Nsoesie(2022)]%
        {ghassemi2022medicine}
\bibfield{author}{\bibinfo{person}{Marzyeh Ghassemi} {and} \bibinfo{person}{Elaine~Okanyene Nsoesie}.} \bibinfo{year}{2022}\natexlab{}.
\newblock \showarticletitle{In medicine, how do we machine learn anything real?}
\newblock \bibinfo{journal}{\emph{Patterns}} (\bibinfo{year}{2022}).
\newblock


\bibitem[Gianfrancesco et~al\mbox{.}(2018)]%
        {gianfrancesco2018potential}
\bibfield{author}{\bibinfo{person}{Milena~A Gianfrancesco}, \bibinfo{person}{Suzanne Tamang}, \bibinfo{person}{Jinoos Yazdany}, {and} \bibinfo{person}{Gabriela Schmajuk}.} \bibinfo{year}{2018}\natexlab{}.
\newblock \showarticletitle{Potential biases in machine learning algorithms using electronic health record data}.
\newblock \bibinfo{journal}{\emph{JAMA internal medicine}} (\bibinfo{year}{2018}).
\newblock


\bibitem[Hahn and Bentley(2003)]%
        {hahn2003drift}
\bibfield{author}{\bibinfo{person}{Matthew~W Hahn} {and} \bibinfo{person}{R~Alexander Bentley}.} \bibinfo{year}{2003}\natexlab{}.
\newblock \showarticletitle{Drift as a mechanism for cultural change: an example from baby names}.
\newblock \bibinfo{journal}{\emph{Proceedings of the Royal Society of London. Series B: Biological Sciences}} (\bibinfo{year}{2003}).
\newblock


\bibitem[Hall et~al\mbox{.}(2015)]%
        {hall2015implicit}
\bibfield{author}{\bibinfo{person}{William~J Hall}, \bibinfo{person}{Mimi~V Chapman}, \bibinfo{person}{Kent~M Lee}, \bibinfo{person}{Yesenia~M Merino}, \bibinfo{person}{Tainayah~W Thomas}, \bibinfo{person}{B~Keith Payne}, \bibinfo{person}{Eugenia Eng}, \bibinfo{person}{Steven~H Day}, {and} \bibinfo{person}{Tamera Coyne-Beasley}.} \bibinfo{year}{2015}\natexlab{}.
\newblock \showarticletitle{Implicit racial/ethnic bias among health care professionals and its influence on health care outcomes: a systematic review}.
\newblock \bibinfo{journal}{\emph{American journal of public health}} (\bibinfo{year}{2015}).
\newblock


\bibitem[Hann{\'a}k et~al\mbox{.}(2017)]%
        {hannak2017bias}
\bibfield{author}{\bibinfo{person}{Anik{\'o} Hann{\'a}k}, \bibinfo{person}{Claudia Wagner}, \bibinfo{person}{David Garcia}, \bibinfo{person}{Alan Mislove}, \bibinfo{person}{Markus Strohmaier}, {and} \bibinfo{person}{Christo Wilson}.} \bibinfo{year}{2017}\natexlab{}.
\newblock \showarticletitle{Bias in online freelance marketplaces: Evidence from taskrabbit and fiverr}. In \bibinfo{booktitle}{\emph{Proceedings of the 2017 ACM conference on computer supported cooperative work and social computing}}.
\newblock


\bibitem[Hanson et~al\mbox{.}(2016)]%
        {hanson2016discrimination}
\bibfield{author}{\bibinfo{person}{Andrew Hanson}, \bibinfo{person}{Zackary Hawley}, \bibinfo{person}{Hal Martin}, {and} \bibinfo{person}{Bo Liu}.} \bibinfo{year}{2016}\natexlab{}.
\newblock \showarticletitle{Discrimination in mortgage lending: Evidence from a correspondence experiment}.
\newblock \bibinfo{journal}{\emph{Journal of Urban Economics}} (\bibinfo{year}{2016}).
\newblock


\bibitem[Harris(2015)]%
        {harris2015s}
\bibfield{author}{\bibinfo{person}{J~Andrew Harris}.} \bibinfo{year}{2015}\natexlab{}.
\newblock \showarticletitle{What's in a name? A method for extracting information about ethnicity from names}.
\newblock \bibinfo{journal}{\emph{Political Analysis}} (\bibinfo{year}{2015}).
\newblock


\bibitem[Hochreiter and Schmidhuber(1997)]%
        {hochreiter1997long}
\bibfield{author}{\bibinfo{person}{Sepp Hochreiter} {and} \bibinfo{person}{J{\"u}rgen Schmidhuber}.} \bibinfo{year}{1997}\natexlab{}.
\newblock \showarticletitle{Long short-term memory}.
\newblock \bibinfo{journal}{\emph{Neural computation}} (\bibinfo{year}{1997}).
\newblock


\bibitem[Honnibal et~al\mbox{.}(2020)]%
        {honnibal2020spacy}
\bibfield{author}{\bibinfo{person}{Matthew Honnibal}, \bibinfo{person}{Ines Montani}, \bibinfo{person}{Sofie Van~Landeghem}, {and} \bibinfo{person}{Adriane Boyd}.} \bibinfo{year}{2020}\natexlab{}.
\newblock \showarticletitle{spaCy: Industrial-strength Natural Language Processing in Python}.
\newblock  (\bibinfo{year}{2020}).
\newblock


\bibitem[Huang et~al\mbox{.}(2020)]%
        {huang2020reducing}
\bibfield{author}{\bibinfo{person}{Po-Sen Huang}, \bibinfo{person}{Huan Zhang}, \bibinfo{person}{Ray Jiang}, \bibinfo{person}{Robert Stanforth}, \bibinfo{person}{Johannes Welbl}, \bibinfo{person}{Jack Rae}, \bibinfo{person}{Vishal Maini}, \bibinfo{person}{Dani Yogatama}, {and} \bibinfo{person}{Pushmeet Kohli}.} \bibinfo{year}{2020}\natexlab{}.
\newblock \showarticletitle{Reducing Sentiment Bias in Language Models via Counterfactual Evaluation}. In \bibinfo{booktitle}{\emph{Findings of the Association for Computational Linguistics: EMNLP 2020}}.
\newblock


\bibitem[Hut-Mossel et~al\mbox{.}(2021)]%
        {hut2021understanding}
\bibfield{author}{\bibinfo{person}{Lisanne Hut-Mossel}, \bibinfo{person}{Kees Ahaus}, \bibinfo{person}{Gera Welker}, {and} \bibinfo{person}{Rijk Gans}.} \bibinfo{year}{2021}\natexlab{}.
\newblock \showarticletitle{Understanding how and why audits work in improving the quality of hospital care: A systematic realist review}.
\newblock \bibinfo{journal}{\emph{PloS one}} (\bibinfo{year}{2021}).
\newblock


\bibitem[Hutchinson and Mitchell(2019)]%
        {hutchinson201950}
\bibfield{author}{\bibinfo{person}{Ben Hutchinson} {and} \bibinfo{person}{Margaret Mitchell}.} \bibinfo{year}{2019}\natexlab{}.
\newblock \showarticletitle{50 years of test (un) fairness: Lessons for machine learning}. In \bibinfo{booktitle}{\emph{Proceedings of the conference on fairness, accountability, and transparency}}.
\newblock


\bibitem[Hutchinson et~al\mbox{.}(2020)]%
        {hutchinson2020social}
\bibfield{author}{\bibinfo{person}{Ben Hutchinson}, \bibinfo{person}{Vinodkumar Prabhakaran}, \bibinfo{person}{Emily Denton}, \bibinfo{person}{Kellie Webster}, \bibinfo{person}{Yu Zhong}, {and} \bibinfo{person}{Stephen Denuyl}.} \bibinfo{year}{2020}\natexlab{}.
\newblock \showarticletitle{Social Biases in NLP Models as Barriers for Persons with Disabilities}. In \bibinfo{booktitle}{\emph{Proceedings of the 58th Annual Meeting of the Association for Computational Linguistics}}.
\newblock


\bibitem[Ivers et~al\mbox{.}(2012)]%
        {ivers2012audit}
\bibfield{author}{\bibinfo{person}{Noah Ivers}, \bibinfo{person}{Gro Jamtvedt}, \bibinfo{person}{Signe Flottorp}, \bibinfo{person}{Jane~M Young}, \bibinfo{person}{Jan Odgaard-Jensen}, \bibinfo{person}{Simon~D French}, \bibinfo{person}{Mary~Ann O'Brien}, \bibinfo{person}{Marit Johansen}, \bibinfo{person}{Jeremy Grimshaw}, {and} \bibinfo{person}{Andrew~D Oxman}.} \bibinfo{year}{2012}\natexlab{}.
\newblock \showarticletitle{Audit and feedback: effects on professional practice and healthcare outcomes}.
\newblock \bibinfo{journal}{\emph{Cochrane database of systematic reviews}} (\bibinfo{year}{2012}).
\newblock


\bibitem[Jacobs et~al\mbox{.}(2020)]%
        {jacobs2020meaning}
\bibfield{author}{\bibinfo{person}{Abigail~Z Jacobs}, \bibinfo{person}{Su~Lin Blodgett}, \bibinfo{person}{Solon Barocas}, \bibinfo{person}{Hal Daum{\'e}~III}, {and} \bibinfo{person}{Hanna Wallach}.} \bibinfo{year}{2020}\natexlab{}.
\newblock \showarticletitle{The meaning and measurement of bias: lessons from natural language processing}. In \bibinfo{booktitle}{\emph{Proceedings of the 2020 Conference on Fairness, Accountability, and Transparency}}.
\newblock


\bibitem[Johnson et~al\mbox{.}(2023)]%
        {johnson2023mimic}
\bibfield{author}{\bibinfo{person}{Alistair~EW Johnson}, \bibinfo{person}{Lucas Bulgarelli}, \bibinfo{person}{Lu Shen}, \bibinfo{person}{Alvin Gayles}, \bibinfo{person}{Ayad Shammout}, \bibinfo{person}{Steven Horng}, \bibinfo{person}{Tom~J Pollard}, \bibinfo{person}{Benjamin Moody}, \bibinfo{person}{Brian Gow}, \bibinfo{person}{Li-wei~H Lehman}, {et~al\mbox{.}}} \bibinfo{year}{2023}\natexlab{}.
\newblock \showarticletitle{MIMIC-IV, a freely accessible electronic health record dataset}.
\newblock \bibinfo{journal}{\emph{Scientific data}} (\bibinfo{year}{2023}).
\newblock


\bibitem[Johnson et~al\mbox{.}(2016)]%
        {johnson2016mimic}
\bibfield{author}{\bibinfo{person}{Alistair~EW Johnson}, \bibinfo{person}{Tom~J Pollard}, \bibinfo{person}{Lu Shen}, \bibinfo{person}{Li-wei~H Lehman}, \bibinfo{person}{Mengling Feng}, \bibinfo{person}{Mohammad Ghassemi}, \bibinfo{person}{Benjamin Moody}, \bibinfo{person}{Peter Szolovits}, \bibinfo{person}{Leo Anthony~Celi}, {and} \bibinfo{person}{Roger~G Mark}.} \bibinfo{year}{2016}\natexlab{}.
\newblock \showarticletitle{MIMIC-III, a freely accessible critical care database}.
\newblock \bibinfo{journal}{\emph{Scientific data}} (\bibinfo{year}{2016}).
\newblock


\bibitem[Kayaalp(2017)]%
        {kayaalp2017modes}
\bibfield{author}{\bibinfo{person}{Mehmet Kayaalp}.} \bibinfo{year}{2017}\natexlab{}.
\newblock \showarticletitle{Modes of De-identification}. In \bibinfo{booktitle}{\emph{AMIA Annual Symposium Proceedings}}.
\newblock


\bibitem[Kiritchenko and Mohammad(2018)]%
        {kiritchenko2018examining}
\bibfield{author}{\bibinfo{person}{Svetlana Kiritchenko} {and} \bibinfo{person}{Saif Mohammad}.} \bibinfo{year}{2018}\natexlab{}.
\newblock \showarticletitle{Examining Gender and Race Bias in Two Hundred Sentiment Analysis Systems}. In \bibinfo{booktitle}{\emph{Proceedings of the Seventh Joint Conference on Lexical and Computational Semantics}}.
\newblock


\bibitem[Kurita et~al\mbox{.}(2019)]%
        {kurita2019measuring}
\bibfield{author}{\bibinfo{person}{Keita Kurita}, \bibinfo{person}{Nidhi Vyas}, \bibinfo{person}{Ayush Pareek}, \bibinfo{person}{Alan~W Black}, {and} \bibinfo{person}{Yulia Tsvetkov}.} \bibinfo{year}{2019}\natexlab{}.
\newblock \showarticletitle{Measuring Bias in Contextualized Word Representations}. In \bibinfo{booktitle}{\emph{Proceedings of the First Workshop on Gender Bias in Natural Language Processing}}.
\newblock


\bibitem[Lafferty et~al\mbox{.}(2001)]%
        {lafferty2001conditional}
\bibfield{author}{\bibinfo{person}{John~D Lafferty}, \bibinfo{person}{Andrew McCallum}, {and} \bibinfo{person}{Fernando~CN Pereira}.} \bibinfo{year}{2001}\natexlab{}.
\newblock \showarticletitle{Conditional Random Fields: Probabilistic Models for Segmenting and Labeling Sequence Data}. In \bibinfo{booktitle}{\emph{Proceedings of the Eighteenth International Conference on Machine Learning}}.
\newblock


\bibitem[Lehman et~al\mbox{.}(2021)]%
        {lehman2021does}
\bibfield{author}{\bibinfo{person}{Eric Lehman}, \bibinfo{person}{Sarthak Jain}, \bibinfo{person}{Karl Pichotta}, \bibinfo{person}{Yoav Goldberg}, {and} \bibinfo{person}{Byron~C Wallace}.} \bibinfo{year}{2021}\natexlab{}.
\newblock \showarticletitle{Does BERT Pretrained on Clinical Notes Reveal Sensitive Data?}. In \bibinfo{booktitle}{\emph{Proceedings of the 2021 Conference of the North American Chapter of the Association for Computational Linguistics: Human Language Technologies}}.
\newblock


\bibitem[Li et~al\mbox{.}(2020)]%
        {li2020survey}
\bibfield{author}{\bibinfo{person}{Jing Li}, \bibinfo{person}{Aixin Sun}, \bibinfo{person}{Jianglei Han}, {and} \bibinfo{person}{Chenliang Li}.} \bibinfo{year}{2020}\natexlab{}.
\newblock \showarticletitle{A survey on deep learning for named entity recognition}.
\newblock \bibinfo{journal}{\emph{IEEE Transactions on Knowledge and Data Engineering}} (\bibinfo{year}{2020}).
\newblock


\bibitem[Liang et~al\mbox{.}(2021)]%
        {liang2021towards}
\bibfield{author}{\bibinfo{person}{Paul~Pu Liang}, \bibinfo{person}{Chiyu Wu}, \bibinfo{person}{Louis-Philippe Morency}, {and} \bibinfo{person}{Ruslan Salakhutdinov}.} \bibinfo{year}{2021}\natexlab{}.
\newblock \showarticletitle{Towards understanding and mitigating social biases in language models}. In \bibinfo{booktitle}{\emph{International Conference on Machine Learning}}.
\newblock


\bibitem[Lim et~al\mbox{.}(2023)]%
        {lim2023annotated}
\bibfield{author}{\bibinfo{person}{Shulammite Lim}, \bibinfo{person}{Alistair Johnson}, \bibinfo{person}{Yuxin Xiao}, \bibinfo{person}{Dana Moukheiber}, \bibinfo{person}{Lama Moukheiber}, \bibinfo{person}{Mira Moukheiber}, \bibinfo{person}{Marzyeh Ghassemi}, {and} \bibinfo{person}{Tom Pollard}.} \bibinfo{year}{2023}\natexlab{}.
\newblock \showarticletitle{Annotated MIMIC-IV discharge summaries for a study on deidentification of names (version 1.0)}.
\newblock \bibinfo{journal}{\emph{PhysioNet}} (\bibinfo{year}{2023}).
\newblock
\newblock
\shownote{https://doi.org/10.13026/ngc0-0f54}.


\bibitem[Liu and Ruths(2013)]%
        {liu2013s}
\bibfield{author}{\bibinfo{person}{Wendy Liu} {and} \bibinfo{person}{Derek Ruths}.} \bibinfo{year}{2013}\natexlab{}.
\newblock \showarticletitle{What’s in a name? using first names as features for gender inference in twitter}. In \bibinfo{booktitle}{\emph{2013 AAAI Spring Symposium Series}}.
\newblock


\bibitem[Liu et~al\mbox{.}(2019)]%
        {liu2019roberta}
\bibfield{author}{\bibinfo{person}{Yinhan Liu}, \bibinfo{person}{Myle Ott}, \bibinfo{person}{Naman Goyal}, \bibinfo{person}{Jingfei Du}, \bibinfo{person}{Mandar Joshi}, \bibinfo{person}{Danqi Chen}, \bibinfo{person}{Omer Levy}, \bibinfo{person}{Mike Lewis}, \bibinfo{person}{Luke Zettlemoyer}, {and} \bibinfo{person}{Veselin Stoyanov}.} \bibinfo{year}{2019}\natexlab{}.
\newblock \showarticletitle{Roberta: A robustly optimized bert pretraining approach}.
\newblock \bibinfo{journal}{\emph{arXiv preprint arXiv:1907.11692}} (\bibinfo{year}{2019}).
\newblock


\bibitem[Lo(2015)]%
        {lo2015sharing}
\bibfield{author}{\bibinfo{person}{Bernard Lo}.} \bibinfo{year}{2015}\natexlab{}.
\newblock \showarticletitle{Sharing clinical trial data: maximizing benefits, minimizing risk}.
\newblock \bibinfo{journal}{\emph{Jama}} (\bibinfo{year}{2015}).
\newblock


\bibitem[Lockhart et~al\mbox{.}(2023)]%
        {lockhart2023name}
\bibfield{author}{\bibinfo{person}{Jeffrey~W Lockhart}, \bibinfo{person}{Molly~M King}, {and} \bibinfo{person}{Christin Munsch}.} \bibinfo{year}{2023}\natexlab{}.
\newblock \showarticletitle{Name-based demographic inference and the unequal distribution of misrecognition}.
\newblock \bibinfo{journal}{\emph{Nature Human Behaviour}} (\bibinfo{year}{2023}).
\newblock


\bibitem[Madan et~al\mbox{.}(2022)]%
        {Madan2022when}
\bibfield{author}{\bibinfo{person}{Spandan Madan}, \bibinfo{person}{Timothy Henry}, \bibinfo{person}{Jamell Dozier}, \bibinfo{person}{Helen Ho}, \bibinfo{person}{Nishchal Bhandari}, \bibinfo{person}{Tomotake Sasaki}, \bibinfo{person}{Fr{\'e}do Durand}, \bibinfo{person}{Hanspeter Pfister}, {and} \bibinfo{person}{Xavier Boix}.} \bibinfo{year}{2022}\natexlab{}.
\newblock \showarticletitle{When and how convolutional neural networks generalize to out-of-distribution category--viewpoint combinations}.
\newblock \bibinfo{journal}{\emph{Nature Machine Intelligence}} (\bibinfo{year}{2022}).
\newblock


\bibitem[Manning et~al\mbox{.}(2014)]%
        {manning2014stanford}
\bibfield{author}{\bibinfo{person}{Christopher~D Manning}, \bibinfo{person}{Mihai Surdeanu}, \bibinfo{person}{John Bauer}, \bibinfo{person}{Jenny~Rose Finkel}, \bibinfo{person}{Steven Bethard}, {and} \bibinfo{person}{David McClosky}.} \bibinfo{year}{2014}\natexlab{}.
\newblock \showarticletitle{The Stanford CoreNLP natural language processing toolkit}. In \bibinfo{booktitle}{\emph{Proceedings of 52nd annual meeting of the association for computational linguistics: system demonstrations}}.
\newblock


\bibitem[Mansfield et~al\mbox{.}(2022)]%
        {mansfield2022behind}
\bibfield{author}{\bibinfo{person}{Courtney Mansfield}, \bibinfo{person}{Amandalynne Paullada}, {and} \bibinfo{person}{Kristen Howell}.} \bibinfo{year}{2022}\natexlab{}.
\newblock \showarticletitle{Behind the Mask: Demographic bias in name detection for PII masking}. In \bibinfo{booktitle}{\emph{Proceedings of the Second Workshop on Language Technology for Equality, Diversity and Inclusion}}.
\newblock


\bibitem[Marcelin et~al\mbox{.}(2019)]%
        {marcelin2019impact}
\bibfield{author}{\bibinfo{person}{Jasmine~R Marcelin}, \bibinfo{person}{Dawd~S Siraj}, \bibinfo{person}{Robert Victor}, \bibinfo{person}{Shaila Kotadia}, {and} \bibinfo{person}{Yvonne~A Maldonado}.} \bibinfo{year}{2019}\natexlab{}.
\newblock \showarticletitle{The impact of unconscious bias in healthcare: how to recognize and mitigate it}.
\newblock \bibinfo{journal}{\emph{The Journal of infectious diseases}} (\bibinfo{year}{2019}).
\newblock


\bibitem[Maudslay et~al\mbox{.}(2019)]%
        {maudslay2019s}
\bibfield{author}{\bibinfo{person}{Rowan~Hall Maudslay}, \bibinfo{person}{Hila Gonen}, \bibinfo{person}{Ryan Cotterell}, {and} \bibinfo{person}{Simone Teufel}.} \bibinfo{year}{2019}\natexlab{}.
\newblock \showarticletitle{It’s All in the Name: Mitigating Gender Bias with Name-Based Counterfactual Data Substitution}. In \bibinfo{booktitle}{\emph{Proceedings of the 2019 Conference on Empirical Methods in Natural Language Processing and the 9th International Joint Conference on Natural Language Processing (EMNLP-IJCNLP)}}.
\newblock


\bibitem[McDermott et~al\mbox{.}(2021)]%
        {mcdermott2021reproducibility}
\bibfield{author}{\bibinfo{person}{Matthew~BA McDermott}, \bibinfo{person}{Shirly Wang}, \bibinfo{person}{Nikki Marinsek}, \bibinfo{person}{Rajesh Ranganath}, \bibinfo{person}{Luca Foschini}, {and} \bibinfo{person}{Marzyeh Ghassemi}.} \bibinfo{year}{2021}\natexlab{}.
\newblock \showarticletitle{Reproducibility in machine learning for health research: Still a ways to go}.
\newblock \bibinfo{journal}{\emph{Science Translational Medicine}} (\bibinfo{year}{2021}).
\newblock


\bibitem[Mehrabi et~al\mbox{.}(2020)]%
        {mehrabi2020man}
\bibfield{author}{\bibinfo{person}{Ninareh Mehrabi}, \bibinfo{person}{Thamme Gowda}, \bibinfo{person}{Fred Morstatter}, \bibinfo{person}{Nanyun Peng}, {and} \bibinfo{person}{Aram Galstyan}.} \bibinfo{year}{2020}\natexlab{}.
\newblock \showarticletitle{Man is to person as woman is to location: Measuring gender bias in named entity recognition}. In \bibinfo{booktitle}{\emph{Proceedings of the 31st ACM Conference on Hypertext and Social Media}}.
\newblock


\bibitem[Mehrabi et~al\mbox{.}(2021)]%
        {mehrabi2021survey}
\bibfield{author}{\bibinfo{person}{Ninareh Mehrabi}, \bibinfo{person}{Fred Morstatter}, \bibinfo{person}{Nripsuta Saxena}, \bibinfo{person}{Kristina Lerman}, {and} \bibinfo{person}{Aram Galstyan}.} \bibinfo{year}{2021}\natexlab{}.
\newblock \showarticletitle{A survey on bias and fairness in machine learning}.
\newblock \bibinfo{journal}{\emph{ACM Computing Surveys (CSUR)}} (\bibinfo{year}{2021}).
\newblock


\bibitem[Meystre et~al\mbox{.}(2010)]%
        {meystre2010automatic}
\bibfield{author}{\bibinfo{person}{Stephane~M Meystre}, \bibinfo{person}{F~Jeffrey Friedlin}, \bibinfo{person}{Brett~R South}, \bibinfo{person}{Shuying Shen}, {and} \bibinfo{person}{Matthew~H Samore}.} \bibinfo{year}{2010}\natexlab{}.
\newblock \showarticletitle{Automatic de-identification of textual documents in the electronic health record: a review of recent research}.
\newblock \bibinfo{journal}{\emph{BMC medical research methodology}} (\bibinfo{year}{2010}).
\newblock


\bibitem[Mishra et~al\mbox{.}(2020)]%
        {mishra2020assessing}
\bibfield{author}{\bibinfo{person}{Shubhanshu Mishra}, \bibinfo{person}{Sijun He}, {and} \bibinfo{person}{Luca Belli}.} \bibinfo{year}{2020}\natexlab{}.
\newblock \showarticletitle{Assessing demographic bias in named entity recognition}.
\newblock \bibinfo{journal}{\emph{arXiv preprint arXiv:2008.03415}} (\bibinfo{year}{2020}).
\newblock


\bibitem[Nadeem et~al\mbox{.}(2021)]%
        {nadeem2021stereoset}
\bibfield{author}{\bibinfo{person}{Moin Nadeem}, \bibinfo{person}{Anna Bethke}, {and} \bibinfo{person}{Siva Reddy}.} \bibinfo{year}{2021}\natexlab{}.
\newblock \showarticletitle{StereoSet: Measuring stereotypical bias in pretrained language models}. In \bibinfo{booktitle}{\emph{Proceedings of the 59th Annual Meeting of the Association for Computational Linguistics and the 11th International Joint Conference on Natural Language Processing (Volume 1: Long Papers)}}.
\newblock


\bibitem[Nangia et~al\mbox{.}(2020)]%
        {nangia2020crows}
\bibfield{author}{\bibinfo{person}{Nikita Nangia}, \bibinfo{person}{Clara Vania}, \bibinfo{person}{Rasika Bhalerao}, {and} \bibinfo{person}{Samuel Bowman}.} \bibinfo{year}{2020}\natexlab{}.
\newblock \showarticletitle{CrowS-Pairs: A Challenge Dataset for Measuring Social Biases in Masked Language Models}. In \bibinfo{booktitle}{\emph{Proceedings of the 2020 Conference on Empirical Methods in Natural Language Processing (EMNLP)}}.
\newblock


\bibitem[Norgeot et~al\mbox{.}(2020)]%
        {norgeot2020protected}
\bibfield{author}{\bibinfo{person}{Beau Norgeot}, \bibinfo{person}{Kathleen Muenzen}, \bibinfo{person}{Thomas~A Peterson}, \bibinfo{person}{Xuancheng Fan}, \bibinfo{person}{Benjamin~S Glicksberg}, \bibinfo{person}{Gundolf Schenk}, \bibinfo{person}{Eugenia Rutenberg}, \bibinfo{person}{Boris Oskotsky}, \bibinfo{person}{Marina Sirota}, \bibinfo{person}{Jinoos Yazdany}, {et~al\mbox{.}}} \bibinfo{year}{2020}\natexlab{}.
\newblock \showarticletitle{Protected Health Information filter (Philter): accurately and securely de-identifying free-text clinical notes}.
\newblock \bibinfo{journal}{\emph{NPJ digital medicine}} (\bibinfo{year}{2020}).
\newblock


\bibitem[Norori et~al\mbox{.}(2021)]%
        {norori2021addressing}
\bibfield{author}{\bibinfo{person}{Natalia Norori}, \bibinfo{person}{Qiyang Hu}, \bibinfo{person}{Florence~Marcelle Aellen}, \bibinfo{person}{Francesca~Dalia Faraci}, {and} \bibinfo{person}{Athina Tzovara}.} \bibinfo{year}{2021}\natexlab{}.
\newblock \showarticletitle{Addressing bias in big data and AI for health care: A call for open science}.
\newblock \bibinfo{journal}{\emph{Patterns}} (\bibinfo{year}{2021}).
\newblock


\bibitem[Ochs(2022)]%
        {ochs2022addressing}
\bibfield{author}{\bibinfo{person}{Jessica~H Ochs}.} \bibinfo{year}{2022}\natexlab{}.
\newblock \showarticletitle{Addressing health disparities by addressing structural racism and implicit bias in nursing education.}
\newblock \bibinfo{journal}{\emph{Nurse Education Today}} (\bibinfo{year}{2022}).
\newblock


\bibitem[OpenAI(2023)]%
        {OpenAI2023GPT4TR}
\bibfield{author}{\bibinfo{person}{OpenAI}.} \bibinfo{year}{2023}\natexlab{}.
\newblock \showarticletitle{GPT-4 Technical Report}.
\newblock \bibinfo{journal}{\emph{arXiv preprint arXiv:2303.08774}} (\bibinfo{year}{2023}).
\newblock


\bibitem[Papakyriakopoulos et~al\mbox{.}(2020)]%
        {papakyriakopoulos2020bias}
\bibfield{author}{\bibinfo{person}{Orestis Papakyriakopoulos}, \bibinfo{person}{Simon Hegelich}, \bibinfo{person}{Juan Carlos~Medina Serrano}, {and} \bibinfo{person}{Fabienne Marco}.} \bibinfo{year}{2020}\natexlab{}.
\newblock \showarticletitle{Bias in word embeddings}. In \bibinfo{booktitle}{\emph{Proceedings of the 2020 conference on fairness, accountability, and transparency}}.
\newblock


\bibitem[Parikh et~al\mbox{.}(2019)]%
        {parikh2019addressing}
\bibfield{author}{\bibinfo{person}{Ravi~B Parikh}, \bibinfo{person}{Stephanie Teeple}, {and} \bibinfo{person}{Amol~S Navathe}.} \bibinfo{year}{2019}\natexlab{}.
\newblock \showarticletitle{Addressing bias in artificial intelligence in health care}.
\newblock \bibinfo{journal}{\emph{Jama}} (\bibinfo{year}{2019}).
\newblock


\bibitem[Pennington et~al\mbox{.}(2014)]%
        {pennington2014glove}
\bibfield{author}{\bibinfo{person}{Jeffrey Pennington}, \bibinfo{person}{Richard Socher}, {and} \bibinfo{person}{Christopher~D Manning}.} \bibinfo{year}{2014}\natexlab{}.
\newblock \showarticletitle{Glove: Global vectors for word representation}. In \bibinfo{booktitle}{\emph{Proceedings of the 2014 conference on empirical methods in natural language processing (EMNLP)}}.
\newblock


\bibitem[Prost et~al\mbox{.}(2019)]%
        {prost2019debiasing}
\bibfield{author}{\bibinfo{person}{Flavien Prost}, \bibinfo{person}{Nithum Thain}, {and} \bibinfo{person}{Tolga Bolukbasi}.} \bibinfo{year}{2019}\natexlab{}.
\newblock \showarticletitle{Debiasing Embeddings for Reduced Gender Bias in Text Classification}. In \bibinfo{booktitle}{\emph{Proceedings of the First Workshop on Gender Bias in Natural Language Processing}}.
\newblock


\bibitem[Qayyum et~al\mbox{.}(2020)]%
        {qayyum2020secure}
\bibfield{author}{\bibinfo{person}{Adnan Qayyum}, \bibinfo{person}{Junaid Qadir}, \bibinfo{person}{Muhammad Bilal}, {and} \bibinfo{person}{Ala Al-Fuqaha}.} \bibinfo{year}{2020}\natexlab{}.
\newblock \showarticletitle{Secure and robust machine learning for healthcare: A survey}.
\newblock \bibinfo{journal}{\emph{IEEE Reviews in Biomedical Engineering}} (\bibinfo{year}{2020}).
\newblock


\bibitem[Qi et~al\mbox{.}(2020)]%
        {qi2020stanza}
\bibfield{author}{\bibinfo{person}{Peng Qi}, \bibinfo{person}{Yuhao Zhang}, \bibinfo{person}{Yuhui Zhang}, \bibinfo{person}{Jason Bolton}, {and} \bibinfo{person}{Christopher~D. Manning}.} \bibinfo{year}{2020}\natexlab{}.
\newblock \showarticletitle{Stanza: A {Python} Natural Language Processing Toolkit for Many Human Languages}. In \bibinfo{booktitle}{\emph{Proceedings of the 58th Annual Meeting of the Association for Computational Linguistics: System Demonstrations}}.
\newblock


\bibitem[Raji and Buolamwini(2019)]%
        {raji2019actionable}
\bibfield{author}{\bibinfo{person}{Inioluwa~Deborah Raji} {and} \bibinfo{person}{Joy Buolamwini}.} \bibinfo{year}{2019}\natexlab{}.
\newblock \showarticletitle{Actionable auditing: Investigating the impact of publicly naming biased performance results of commercial ai products}. In \bibinfo{booktitle}{\emph{Proceedings of the 2019 AAAI/ACM Conference on AI, Ethics, and Society}}.
\newblock


\bibitem[Rudinger et~al\mbox{.}(2018)]%
        {rudinger2018gender}
\bibfield{author}{\bibinfo{person}{Rachel Rudinger}, \bibinfo{person}{Jason Naradowsky}, \bibinfo{person}{Brian Leonard}, {and} \bibinfo{person}{Benjamin Van~Durme}.} \bibinfo{year}{2018}\natexlab{}.
\newblock \showarticletitle{Gender Bias in Coreference Resolution}. In \bibinfo{booktitle}{\emph{Proceedings of the 2018 Conference of the North American Chapter of the Association for Computational Linguistics: Human Language Technologies, Volume 2 (Short Papers)}}.
\newblock


\bibitem[Sang and De~Meulder(2003)]%
        {sang2003introduction}
\bibfield{author}{\bibinfo{person}{Erik Tjong~Kim Sang} {and} \bibinfo{person}{Fien De~Meulder}.} \bibinfo{year}{2003}\natexlab{}.
\newblock \showarticletitle{Introduction to the CoNLL-2003 Shared Task: Language-Independent Named Entity Recognition}. In \bibinfo{booktitle}{\emph{Proceedings of the Seventh Conference on Natural Language Learning at HLT-NAACL 2003}}.
\newblock


\bibitem[Sap et~al\mbox{.}(2019)]%
        {sap2019risk}
\bibfield{author}{\bibinfo{person}{Maarten Sap}, \bibinfo{person}{Dallas Card}, \bibinfo{person}{Saadia Gabriel}, \bibinfo{person}{Yejin Choi}, {and} \bibinfo{person}{Noah~A Smith}.} \bibinfo{year}{2019}\natexlab{}.
\newblock \showarticletitle{The risk of racial bias in hate speech detection}. In \bibinfo{booktitle}{\emph{Proceedings of the 57th annual meeting of the association for computational linguistics}}.
\newblock


\bibitem[Savoldi et~al\mbox{.}(2021)]%
        {savoldi2021gender}
\bibfield{author}{\bibinfo{person}{Beatrice Savoldi}, \bibinfo{person}{Marco Gaido}, \bibinfo{person}{Luisa Bentivogli}, \bibinfo{person}{Matteo Negri}, {and} \bibinfo{person}{Marco Turchi}.} \bibinfo{year}{2021}\natexlab{}.
\newblock \showarticletitle{Gender bias in machine translation}.
\newblock \bibinfo{journal}{\emph{Transactions of the Association for Computational Linguistics}} (\bibinfo{year}{2021}).
\newblock


\bibitem[Schweter and Akbik(2020)]%
        {schweter2020flert}
\bibfield{author}{\bibinfo{person}{Stefan Schweter} {and} \bibinfo{person}{Alan Akbik}.} \bibinfo{year}{2020}\natexlab{}.
\newblock \showarticletitle{Flert: Document-level features for named entity recognition}.
\newblock \bibinfo{journal}{\emph{arXiv preprint arXiv:2011.06993}} (\bibinfo{year}{2020}).
\newblock


\bibitem[Seastedt et~al\mbox{.}(2022)]%
        {seastedt2022global}
\bibfield{author}{\bibinfo{person}{Kenneth~P Seastedt}, \bibinfo{person}{Patrick Schwab}, \bibinfo{person}{Zach O’Brien}, \bibinfo{person}{Edith Wakida}, \bibinfo{person}{Karen Herrera}, \bibinfo{person}{Portia Grace~F Marcelo}, \bibinfo{person}{Louis Agha-Mir-Salim}, \bibinfo{person}{Xavier~Borrat Frigola}, \bibinfo{person}{Emily~Boardman Ndulue}, \bibinfo{person}{Alvin Marcelo}, {et~al\mbox{.}}} \bibinfo{year}{2022}\natexlab{}.
\newblock \showarticletitle{Global healthcare fairness: We should be sharing more, not less, data}.
\newblock \bibinfo{journal}{\emph{PLOS Digital Health}} (\bibinfo{year}{2022}).
\newblock


\bibitem[Seyyed-Kalantari et~al\mbox{.}(2021)]%
        {seyyed2021medical}
\bibfield{author}{\bibinfo{person}{Laleh Seyyed-Kalantari}, \bibinfo{person}{Guanxiong Liu}, \bibinfo{person}{Matthew McDermott}, \bibinfo{person}{Irene Chen}, {and} \bibinfo{person}{Marzyeh Ghassemi}.} \bibinfo{year}{2021}\natexlab{}.
\newblock \showarticletitle{Medical imaging algorithms exacerbate biases in underdiagnosis}.
\newblock  (\bibinfo{year}{2021}).
\newblock


\bibitem[Shah et~al\mbox{.}(2020)]%
        {shah2020predictive}
\bibfield{author}{\bibinfo{person}{Deven~Santosh Shah}, \bibinfo{person}{H~Andrew Schwartz}, {and} \bibinfo{person}{Dirk Hovy}.} \bibinfo{year}{2020}\natexlab{}.
\newblock \showarticletitle{Predictive Biases in Natural Language Processing Models: A Conceptual Framework and Overview}. In \bibinfo{booktitle}{\emph{Proceedings of the 58th Annual Meeting of the Association for Computational Linguistics}}.
\newblock


\bibitem[Shailaja et~al\mbox{.}(2018)]%
        {shailaja2018machine}
\bibfield{author}{\bibinfo{person}{K Shailaja}, \bibinfo{person}{B Seetharamulu}, {and} \bibinfo{person}{MA Jabbar}.} \bibinfo{year}{2018}\natexlab{}.
\newblock \showarticletitle{Machine learning in healthcare: A review}. In \bibinfo{booktitle}{\emph{2018 Second international conference on electronics, communication and aerospace technology (ICECA)}}.
\newblock


\bibitem[Shin et~al\mbox{.}(2020)]%
        {shin2020neutralizing}
\bibfield{author}{\bibinfo{person}{Seungjae Shin}, \bibinfo{person}{Kyungwoo Song}, \bibinfo{person}{JoonHo Jang}, \bibinfo{person}{Hyemi Kim}, \bibinfo{person}{Weonyoung Joo}, {and} \bibinfo{person}{Il-Chul Moon}.} \bibinfo{year}{2020}\natexlab{}.
\newblock \showarticletitle{Neutralizing Gender Bias in Word Embeddings with Latent Disentanglement and Counterfactual Generation}. In \bibinfo{booktitle}{\emph{Findings of the Association for Computational Linguistics: EMNLP 2020}}.
\newblock


\bibitem[Song et~al\mbox{.}(2021)]%
        {song2021deep}
\bibfield{author}{\bibinfo{person}{Bosheng Song}, \bibinfo{person}{Fen Li}, \bibinfo{person}{Yuansheng Liu}, {and} \bibinfo{person}{Xiangxiang Zeng}.} \bibinfo{year}{2021}\natexlab{}.
\newblock \showarticletitle{Deep learning methods for biomedical named entity recognition: a survey and qualitative comparison}.
\newblock \bibinfo{journal}{\emph{Briefings in Bioinformatics}} (\bibinfo{year}{2021}).
\newblock


\bibitem[Stanovsky et~al\mbox{.}(2019)]%
        {stanovsky2019evaluating}
\bibfield{author}{\bibinfo{person}{Gabriel Stanovsky}, \bibinfo{person}{Noah~A Smith}, {and} \bibinfo{person}{Luke Zettlemoyer}.} \bibinfo{year}{2019}\natexlab{}.
\newblock \showarticletitle{Evaluating Gender Bias in Machine Translation}. In \bibinfo{booktitle}{\emph{Proceedings of the 57th Annual Meeting of the Association for Computational Linguistics}}.
\newblock


\bibitem[Stubbs and Uzuner(2015)]%
        {stubbs2015annotating}
\bibfield{author}{\bibinfo{person}{Amber Stubbs} {and} \bibinfo{person}{{\"O}zlem Uzuner}.} \bibinfo{year}{2015}\natexlab{}.
\newblock \showarticletitle{Annotating longitudinal clinical narratives for de-identification: The 2014 i2b2/UTHealth corpus}.
\newblock \bibinfo{journal}{\emph{Journal of biomedical informatics}} (\bibinfo{year}{2015}).
\newblock


\bibitem[Sun et~al\mbox{.}(2019)]%
        {sun2019mitigating}
\bibfield{author}{\bibinfo{person}{Tony Sun}, \bibinfo{person}{Andrew Gaut}, \bibinfo{person}{Shirlyn Tang}, \bibinfo{person}{Yuxin Huang}, \bibinfo{person}{Mai ElSherief}, \bibinfo{person}{Jieyu Zhao}, \bibinfo{person}{Diba Mirza}, \bibinfo{person}{Elizabeth Belding}, \bibinfo{person}{Kai-Wei Chang}, {and} \bibinfo{person}{William~Yang Wang}.} \bibinfo{year}{2019}\natexlab{}.
\newblock \showarticletitle{Mitigating Gender Bias in Natural Language Processing: Literature Review}. In \bibinfo{booktitle}{\emph{Proceedings of the 57th Annual Meeting of the Association for Computational Linguistics}}.
\newblock


\bibitem[Thomas et~al\mbox{.}(2002)]%
        {thomas2002successful}
\bibfield{author}{\bibinfo{person}{Sean~M Thomas}, \bibinfo{person}{Burke Mamlin}, \bibinfo{person}{Gunther Schadow}, {and} \bibinfo{person}{Clement McDonald}.} \bibinfo{year}{2002}\natexlab{}.
\newblock \showarticletitle{A successful technique for removing names in pathology reports using an augmented search and replace method.}. In \bibinfo{booktitle}{\emph{Proceedings of the AMIA Symposium}}.
\newblock


\bibitem[Toma{\v{s}}ev et~al\mbox{.}(2019)]%
        {tomavsev2019clinically}
\bibfield{author}{\bibinfo{person}{Nenad Toma{\v{s}}ev}, \bibinfo{person}{Xavier Glorot}, \bibinfo{person}{Jack~W Rae}, \bibinfo{person}{Michal Zielinski}, \bibinfo{person}{Harry Askham}, \bibinfo{person}{Andre Saraiva}, \bibinfo{person}{Anne Mottram}, \bibinfo{person}{Clemens Meyer}, \bibinfo{person}{Suman Ravuri}, \bibinfo{person}{Ivan Protsyuk}, {et~al\mbox{.}}} \bibinfo{year}{2019}\natexlab{}.
\newblock \showarticletitle{A clinically applicable approach to continuous prediction of future acute kidney injury}.
\newblock \bibinfo{journal}{\emph{Nature}} (\bibinfo{year}{2019}).
\newblock


\bibitem[Topol(2019)]%
        {topol2019high}
\bibfield{author}{\bibinfo{person}{Eric~J Topol}.} \bibinfo{year}{2019}\natexlab{}.
\newblock \showarticletitle{High-performance medicine: the convergence of human and artificial intelligence}.
\newblock \bibinfo{journal}{\emph{Nature medicine}} (\bibinfo{year}{2019}).
\newblock


\bibitem[TSIMA(2023)]%
        {tsima2023reproducibility}
\bibfield{author}{\bibinfo{person}{K TSIMA}.} \bibinfo{year}{2023}\natexlab{}.
\newblock \showarticletitle{The reproducibility issues that haunt health-care AI}.
\newblock \bibinfo{journal}{\emph{Nature}} (\bibinfo{year}{2023}).
\newblock


\bibitem[Tucker et~al\mbox{.}(2016)]%
        {tucker2016protecting}
\bibfield{author}{\bibinfo{person}{Katherine Tucker}, \bibinfo{person}{Janice Branson}, \bibinfo{person}{Maria Dilleen}, \bibinfo{person}{Sally Hollis}, \bibinfo{person}{Paul Loughlin}, \bibinfo{person}{Mark~J Nixon}, {and} \bibinfo{person}{Zo{\"e} Williams}.} \bibinfo{year}{2016}\natexlab{}.
\newblock \showarticletitle{Protecting patient privacy when sharing patient-level data from clinical trials}.
\newblock \bibinfo{journal}{\emph{BMC medical research methodology}} (\bibinfo{year}{2016}).
\newblock


\bibitem[Tzioumis(2018)]%
        {tzioumis2018demographic}
\bibfield{author}{\bibinfo{person}{Konstantinos Tzioumis}.} \bibinfo{year}{2018}\natexlab{}.
\newblock \showarticletitle{Demographic aspects of first names}.
\newblock \bibinfo{journal}{\emph{Scientific data}} (\bibinfo{year}{2018}).
\newblock


\bibitem[Uzuner et~al\mbox{.}(2007)]%
        {uzuner2007evaluating}
\bibfield{author}{\bibinfo{person}{{\"O}zlem Uzuner}, \bibinfo{person}{Yuan Luo}, {and} \bibinfo{person}{Peter Szolovits}.} \bibinfo{year}{2007}\natexlab{}.
\newblock \showarticletitle{Evaluating the state-of-the-art in automatic de-identification}.
\newblock \bibinfo{journal}{\emph{Journal of the American Medical Informatics Association}} (\bibinfo{year}{2007}).
\newblock


\bibitem[Uzuner et~al\mbox{.}(2008)]%
        {uzuner2008identifier}
\bibfield{author}{\bibinfo{person}{{\"O}zlem Uzuner}, \bibinfo{person}{Tawanda~C Sibanda}, \bibinfo{person}{Yuan Luo}, {and} \bibinfo{person}{Peter Szolovits}.} \bibinfo{year}{2008}\natexlab{}.
\newblock \showarticletitle{A de-identifier for medical discharge summaries}.
\newblock \bibinfo{journal}{\emph{Artificial intelligence in medicine}} (\bibinfo{year}{2008}).
\newblock


\bibitem[Webster et~al\mbox{.}(2022)]%
        {webster2022social}
\bibfield{author}{\bibinfo{person}{Craig~S Webster}, \bibinfo{person}{Saana Taylor}, \bibinfo{person}{Courtney Thomas}, {and} \bibinfo{person}{Jennifer~M Weller}.} \bibinfo{year}{2022}\natexlab{}.
\newblock \showarticletitle{Social bias, discrimination and inequity in healthcare: mechanisms, implications and recommendations}.
\newblock \bibinfo{journal}{\emph{BJA education}} (\bibinfo{year}{2022}).
\newblock


\bibitem[Weischedel et~al\mbox{.}(2013)]%
        {weischedel2013ontonotes}
\bibfield{author}{\bibinfo{person}{Ralph Weischedel}, \bibinfo{person}{Martha Palmer}, \bibinfo{person}{Mitchell Marcus}, \bibinfo{person}{Eduard Hovy}, \bibinfo{person}{Sameer Pradhan}, \bibinfo{person}{Lance Ramshaw}, \bibinfo{person}{Nianwen Xue}, \bibinfo{person}{Ann Taylor}, \bibinfo{person}{Jeff Kaufman}, \bibinfo{person}{Michelle Franchini}, {et~al\mbox{.}}} \bibinfo{year}{2013}\natexlab{}.
\newblock \showarticletitle{Ontonotes release 5.0 ldc2013t19}.
\newblock \bibinfo{journal}{\emph{Linguistic Data Consortium, Philadelphia, PA}} (\bibinfo{year}{2013}).
\newblock


\bibitem[Weisstein(2004)]%
        {weisstein2004bonferroni}
\bibfield{author}{\bibinfo{person}{Eric~W Weisstein}.} \bibinfo{year}{2004}\natexlab{}.
\newblock \showarticletitle{Bonferroni correction}.
\newblock \bibinfo{journal}{\emph{https://mathworld. wolfram. com/}} (\bibinfo{year}{2004}).
\newblock


\bibitem[Williams and Wyatt(2015)]%
        {williams2015racial}
\bibfield{author}{\bibinfo{person}{David~R Williams} {and} \bibinfo{person}{Ronald Wyatt}.} \bibinfo{year}{2015}\natexlab{}.
\newblock \showarticletitle{Racial bias in health care and health: challenges and opportunities}.
\newblock \bibinfo{journal}{\emph{JAMA}} (\bibinfo{year}{2015}).
\newblock


\bibitem[Woolson(2007)]%
        {woolson2007wilcoxon}
\bibfield{author}{\bibinfo{person}{Robert~F Woolson}.} \bibinfo{year}{2007}\natexlab{}.
\newblock \showarticletitle{Wilcoxon signed-rank test}.
\newblock \bibinfo{journal}{\emph{Wiley encyclopedia of clinical trials}} (\bibinfo{year}{2007}).
\newblock


\bibitem[Yadav and Bethard(2018)]%
        {yadav2018survey}
\bibfield{author}{\bibinfo{person}{Vikas Yadav} {and} \bibinfo{person}{Steven Bethard}.} \bibinfo{year}{2018}\natexlab{}.
\newblock \showarticletitle{A Survey on Recent Advances in Named Entity Recognition from Deep Learning models}. In \bibinfo{booktitle}{\emph{Proceedings of the 27th International Conference on Computational Linguistics}}.
\newblock


\bibitem[Yang and Garibaldi(2015)]%
        {yang2015automatic}
\bibfield{author}{\bibinfo{person}{Hui Yang} {and} \bibinfo{person}{Jonathan~M Garibaldi}.} \bibinfo{year}{2015}\natexlab{}.
\newblock \showarticletitle{Automatic detection of protected health information from clinic narratives}.
\newblock \bibinfo{journal}{\emph{Journal of biomedical informatics}} (\bibinfo{year}{2015}).
\newblock


\bibitem[Yao et~al\mbox{.}(2019)]%
        {yao2019DocRED}
\bibfield{author}{\bibinfo{person}{Yuan Yao}, \bibinfo{person}{Deming Ye}, \bibinfo{person}{Peng Li}, \bibinfo{person}{Xu Han}, \bibinfo{person}{Yankai Lin}, \bibinfo{person}{Zhenghao Liu}, \bibinfo{person}{Zhiyuan Liu}, \bibinfo{person}{Lixin Huang}, \bibinfo{person}{Jie Zhou}, {and} \bibinfo{person}{Maosong Sun}.} \bibinfo{year}{2019}\natexlab{}.
\newblock \showarticletitle{{DocRED}: A Large-Scale Document-Level Relation Extraction Dataset}. In \bibinfo{booktitle}{\emph{Proceedings of ACL 2019}}.
\newblock


\bibitem[Zestcott et~al\mbox{.}(2016)]%
        {zestcott2016examining}
\bibfield{author}{\bibinfo{person}{Colin~A Zestcott}, \bibinfo{person}{Irene~V Blair}, {and} \bibinfo{person}{Jeff Stone}.} \bibinfo{year}{2016}\natexlab{}.
\newblock \showarticletitle{Examining the presence, consequences, and reduction of implicit bias in health care: a narrative review}.
\newblock \bibinfo{journal}{\emph{Group Processes \& Intergroup Relations}} (\bibinfo{year}{2016}).
\newblock


\bibitem[Zhang et~al\mbox{.}(2020)]%
        {zhang2020hurtful}
\bibfield{author}{\bibinfo{person}{Haoran Zhang}, \bibinfo{person}{Amy~X Lu}, \bibinfo{person}{Mohamed Abdalla}, \bibinfo{person}{Matthew McDermott}, {and} \bibinfo{person}{Marzyeh Ghassemi}.} \bibinfo{year}{2020}\natexlab{}.
\newblock \showarticletitle{Hurtful words: quantifying biases in clinical contextual word embeddings}. In \bibinfo{booktitle}{\emph{proceedings of the ACM Conference on Health, Inference, and Learning}}.
\newblock


\bibitem[Zhao et~al\mbox{.}(2018)]%
        {zhao2018gender}
\bibfield{author}{\bibinfo{person}{Jieyu Zhao}, \bibinfo{person}{Tianlu Wang}, \bibinfo{person}{Mark Yatskar}, \bibinfo{person}{Vicente Ordonez}, {and} \bibinfo{person}{Kai-Wei Chang}.} \bibinfo{year}{2018}\natexlab{}.
\newblock \showarticletitle{Gender Bias in Coreference Resolution: Evaluation and Debiasing Methods}. In \bibinfo{booktitle}{\emph{Proceedings of the 2018 Conference of the North American Chapter of the Association for Computational Linguistics: Human Language Technologies, Volume 2 (Short Papers)}}.
\newblock


\end{thebibliography}

\newpage
\appendix
\onecolumn
\begin{figure*}[t!]
    \centering
    \includegraphics[width=\textwidth]{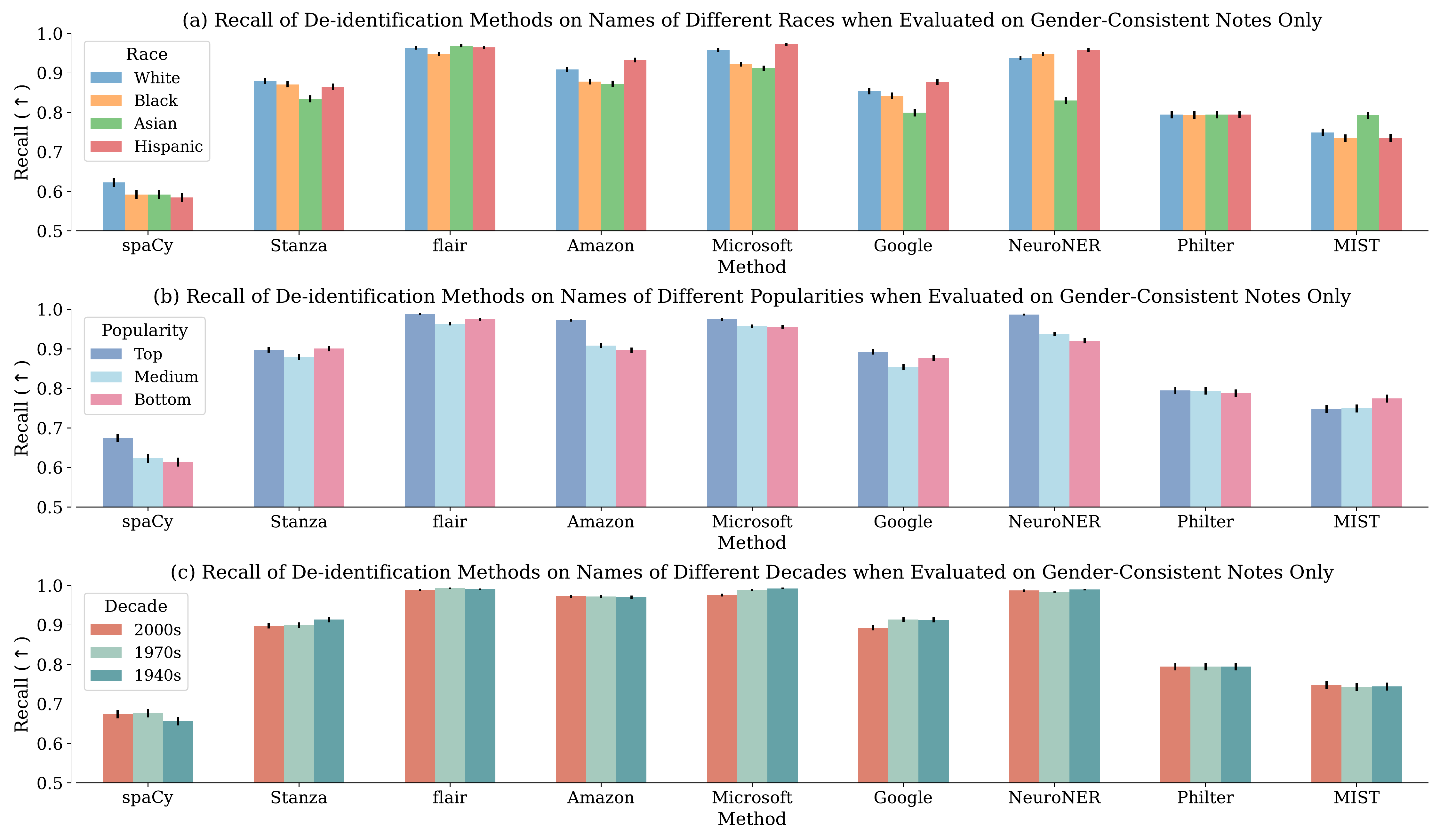}
    \caption{Recall and 95\% bootstrapped confidence interval of the demographic groups along the dimensions of race, name popularity, and the decade of popularity by each examined de-identification method under gender-consistent evaluation. These methods behave similarly compared to the original setup in Figure~\ref{fig:dimension_recall}.}
    \label{fig:dimension_recall_no_gender}
\end{figure*}

\section{Appendix} \label{app:1}
In the appendix, we include additional analysis exploring the robustness of our results in gender-consistent note population, in the subset of notes with the poorest overall performance, and using another fairness metric of recall maximum difference.

\subsection{Gender-Consistent Note Population} \label{app:1.1}
To examine the influence of gender-inconsistent pronouns used in our note template population, we run a robustness check on our results where we only consider male-originating clinical notes populated with male name sets and female-originating notes populated with female name sets. 
We note that in this setting, we do not conduct a direct comparison of the gender gap since the male- and female-originating notes are disjoint.
Otherwise, the experiment follows the procedure in Sec~\ref{sec:3}.

Figure~\ref{fig:dimension_recall_no_gender} illustrates the recall of the demographic groups along the dimensions of race, name popularity, and the decade of popularity by each de-identification method under this gender-confirming evaluation setup. 
The Wilcoxon signed-rank test with p-value $=0.082$ indicates that these methods behave consistently to the original setup in Sec~\ref{sec:3}, and our observations about the race, popularity, and decade disparities based on Figure~\ref{fig:dimension_recall} still hold.
\clearpage

\begin{table*}[t!]
  \centering
    \begin{tabular}{c|ccc|cccc}
    \toprule
        \multirow{2}{*}{\textbf{Method}} & \multicolumn{3}{c|}{\textbf{Overall Performance ($\uparrow$)}} & \multicolumn{4}{c}{\textbf{Bias along Dimensions ($\downarrow$)}} \\
        & \textbf{Precision} & \textbf{Recall} & \textbf{F1} & \textbf{Gender} & \textbf{Race} & \textbf{Popularity} & \textbf{Decade} \\
    \midrule 
    \midrule
        \texttt{spaCy} & 0.874$\pm$0.003 & 0.504$\pm$0.003 & 0.640$\pm$0.003 & 0.004$\pm$0.003 & 0.022$^*\pm$0.004 & 0.037$^*\pm$0.005 & 0.005$\pm$0.004 \\
        \texttt{Stanza} & 0.615$\pm$0.003 & 0.791$\pm$0.003 & 0.692$\pm$0.002 & \textbf{0.001}$\pm$\textbf{0.002} & \textbf{0.007}$\pm$\textbf{0.003} & 0.028$^*\pm$0.004 & 0.011$^*\pm$0.003 \\
        \texttt{flair} & 0.878$\pm$0.002 & \textbf{0.945}$\pm$\textbf{0.001} & \textbf{0.910}$\pm$\textbf{0.001} & 0.005$^*\pm$0.001 & 0.014$^*\pm$0.002 & 0.016$^*\pm$0.002 & 0.004$^*\pm$0.001 \\
        \texttt{Amazon} & \textbf{0.882}$\pm$\textbf{0.002} & 0.883$\pm$0.002 & 0.883$\pm$0.002 & 0.009$^*\pm$0.002 & 0.025$^*\pm$0.003 & 0.047$^*\pm$0.003 & \textbf{0.003}$^*\pm$\textbf{0.002} \\
        \texttt{Microsoft} & 0.619$\pm$0.003 & \textbf{0.936}$\pm$\textbf{0.002} & 0.745$\pm$0.002 & 0.003$^*\pm$0.001 & 0.033$^*\pm$0.003 & 0.013$^*\pm$0.002 & 0.009$^*\pm$0.002 \\
        \texttt{Google} & 0.558$\pm$0.003 & 0.856$\pm$0.002 & 0.676$\pm$0.002 & 0.011$^*\pm$0.002 & 0.034$^*\pm$0.003 & \textbf{0.011}$^*\pm$\textbf{0.003} & 0.008$^*\pm$0.003 \\
        \texttt{NeuroNER} & \textbf{0.929}$\pm$\textbf{0.002} & 0.899$\pm$0.002 & \textbf{0.914}$\pm$\textbf{0.001} & 0.005$^*\pm$0.002 & 0.044$^*\pm$0.003 & 0.052$^*\pm$0.003 & 0.005$\pm$0.002 \\
        \texttt{Philter} & 0.134$\pm$0.001 & 0.562$\pm$0.003 & 0.216$\pm$0.002 & \textbf{0.000}$\pm$\textbf{0.002} & \textbf{0.000}$\pm$\textbf{0.003} & \textbf{0.003}$^*\pm$\textbf{0.004} & \textbf{0.000}$\pm$\textbf{0.004} \\
        \texttt{MIST} & 0.306$\pm$0.002 & 0.532$\pm$0.003 & 0.388$\pm$0.002 & 0.020$^*\pm$0.003 & 0.040$^*\pm$0.004 & 0.019$^*\pm$0.005 & 0.009$^*\pm$0.004 \\
    \bottomrule 
  \end{tabular}
  \caption{Overall performance (higher is better), bias along demographic dimensions (lower is better), and the associated bootstrap standard error of the examined de-identification methods on the hardest 20 templates. We measure the bias with recall equality difference and bold the best two scores in each column. These methods' overall performance follows the general pattern when evaluated on the full set of 100 templates in Table~\ref{tab:overall_performance}. Some methods exhibit lower bias here, possibly due to equally poor performance across demographic groups in harder templates.}
  \label{tab:overall_performance_hardest} 
\end{table*}

\begin{table*}[t!]
  \centering
    \begin{tabular}{c|cccc}
    \toprule
        \multirow{2}{*}{\textbf{Method}} & \multicolumn{4}{c}{\textbf{Recall Maximum Difference ($\downarrow$)}} \\
        & \textbf{Gender} & \textbf{Race} & \textbf{Popularity} & \textbf{Decade} \\
    \midrule 
    \midrule
        \texttt{spaCy} & 0.002$\pm$0.002 & 0.025$\pm$0.004 & 0.042$\pm$0.004 & 0.010$\pm$0.004 \\
        \texttt{Stanza} & 0.002$\pm$0.001 & 0.032$\pm$0.003 & 0.017$\pm$0.003 & 0.008$\pm$0.002 \\
        \texttt{flair} & 0.003$\pm$0.001 & \textbf{0.013}$\pm$\textbf{0.002} & \textbf{0.013}$\pm$\textbf{0.001} & 0.003$\pm$0.001 \\
        \texttt{Amazon} & 0.005$\pm$0.001 & 0.034$\pm$0.002 & 0.047$\pm$0.002 & \textbf{0.001}$\pm$\textbf{0.00}1 \\
        \texttt{Microsoft} & 0.003$\pm$0.001 & 0.033$\pm$0.002 & 0.015$\pm$0.001 & 0.009$\pm$0.001 \\
        \texttt{Google} & 0.009$\pm$0.001 & 0.044$\pm$0.003 & 0.020$\pm$0.003 & 0.015$\pm$0.003 \\
        \texttt{NeuroNER} & \textbf{0.001}$\pm$\textbf{0.001} & 0.089$\pm$0.003 & 0.040$\pm$0.001 & 0.003$\pm$0.001 \\
        \texttt{Philter} & \textbf{0.000}$\pm$\textbf{0.001} & \textbf{0.000}$\pm$\textbf{0.002} & \textbf{0.004}$\pm$\textbf{0.003} & \textbf{0.000}$\pm$\textbf{0.002} \\
        \texttt{MIST} & 0.013$\pm$0.002 & 0.043$\pm$0.004 & 0.026$\pm$0.004 & 0.004$\pm$0.003 \\
    \bottomrule 
  \end{tabular}
  \caption{Recall maximum difference (lower is better) and the associated bootstrapped standard error of the examined de-identification methods. We bold the best two scores in each column. The bias in these methods measured by recall maximum difference along each dimension is similar to the pattern measured by recall equality difference in Table~\ref{tab:overall_performance}.}
  \label{tab:overall_maximum_difference} 
\end{table*}

\subsection{Evaluation of Difficult Note Templates} \label{app:1.2}
Here we identify the set of 20 templates that receive the lowest average recall by the examined de-identification methods and investigate how the performance of these methods changes in Table~\ref{tab:overall_performance_hardest} when evaluated on these harder templates.
Although the scores of their overall performance drop compared to Table~\ref{tab:overall_performance}, the best-performing methods based on the original full set of 100 templates still perform well on these hardest 20 templates.
However, some of the examined methods, such as \texttt{Stanza} and \texttt{Google}, exhibit lower bias now, potentially due to equally poor performance across demographic groups in harder templates.

\subsection{Recall Maximum Difference} \label{app:1.3}
Besides recall equality difference, we consider an additional fairness metric---recall maximum difference, which illustrates the largest gap in recall any demographic group would experience while anticipating the reported average performance.
For dimension $D$ and its entailed set of demographic groups $\mathcal{G}^D = \{\mathcal{G}_1^D, \mathcal{G}_2^D, \dots\}$, recall maximum difference $= \max_{\mathcal{G}_i^D \in \mathcal{G}^D} |\text{Recall}(\mathcal{G}_i^D) - \text{Recall}(\mathcal{G}^D)|$.

Table~\ref{tab:overall_maximum_difference} displays the recall maximum difference of each examined de-identification method along each dimension.
These methods' behaviors here are similar to their bias measured by recall equality difference in Table~\ref{tab:overall_performance}.
Methods that attain the lowest recall equality difference still perform well in terms of recall maximum difference.
\newpage
\begin{table*}[t!]
  \centering
    \begin{tabular}{c|ccc|cccc}
    \toprule
        \multirow{2}{*}{\textbf{Method}} & \multicolumn{3}{c|}{\textbf{Overall Performance ($\uparrow$)}} & \multicolumn{4}{c}{\textbf{Bias along Dimensions ($\downarrow$)}} \\
        & \textbf{Precision} & \textbf{Recall} & \textbf{F1} & \textbf{Gender} & \textbf{Race} & \textbf{Popularity} & \textbf{Decade} \\
    \midrule 
    \midrule
        \texttt{spaCy} & 0.917$\pm$0.001 & 0.629$\pm$0.001 & 0.746$\pm$0.001 & 0.002$^*\pm$0.001 & 0.013$^*\pm$0.002 & 0.028$^*\pm$0.002 & 0.007$^*\pm$0.002 \\
        \texttt{Stanza} & 0.678$\pm$0.001 & 0.881$\pm$0.001 & 0.766$\pm$0.001 & 0.002$^*\pm$0.001 & 0.016$^*\pm$0.002 & 0.011$^*\pm$0.001 & 0.005$^*\pm$0.001 \\
        \texttt{flair} & 0.920$\pm$0.001 & \textbf{0.974}$\pm$\textbf{0.000} & \textbf{0.946}$\pm$\textbf{0.000} & 0.003$^*\pm$0.000 & \textbf{0.006}$^*\pm$\textbf{0.001} & \textbf{0.008}$^*\pm$\textbf{0.001} & 0.002$^*\pm$0.000 \\
        \texttt{Amazon} & \textbf{0.923}$\pm$\textbf{0.001} & 0.925$\pm$0.001 & 0.924$\pm$0.001 & 0.005$^*\pm$0.001 & 0.022$^*\pm$0.001 & 0.032$^*\pm$0.001 & \textbf{0.001}$\pm$\textbf{0.001} \\
        \texttt{Microsoft} & 0.664$\pm$0.001 & \textbf{0.960}$\pm$\textbf{0.001} & 0.785$\pm$0.001 & 0.003$^*\pm$0.001 & 0.023$^*\pm$0.001 & 0.010$^*\pm$0.001 & 0.006$^*\pm$0.001 \\
        \texttt{Google} & 0.609$\pm$0.001 & 0.869$\pm$0.001 & 0.716$\pm$0.001 & 0.009$^*\pm$0.001 & 0.025$^*\pm$0.001 & 0.014$^*\pm$0.002 & 0.010$^*\pm$0.001 \\
        \texttt{NeuroNER} & \textbf{0.946}$\pm$\textbf{0.001} & 0.944$\pm$0.001 & \textbf{0.945}$\pm$\textbf{0.000} & \textbf{0.001}$\pm$\textbf{0.001} & 0.045$^*\pm$0.001 & 0.026$^*\pm$0.001 & 0.002$\pm$0.001 \\
        \texttt{Philter} & 0.227$\pm$0.001 & 0.794$\pm$0.001 & 0.353$\pm$0.001 & \textbf{0.000}$\pm$\textbf{0.001} & \textbf{0.000}$\pm$\textbf{0.001} & \textbf{0.003}$^*\pm$\textbf{0.002} & \textbf{0.000}$\pm$\textbf{0.001} \\
        \texttt{MIST} & 0.474$\pm$0.001 & 0.751$\pm$0.001 & 0.581$\pm$0.001 & 0.013$^*\pm$0.001 & 0.022$^*\pm$0.002 & 0.017$^*\pm$0.002 & 0.003$^*\pm$0.002 \\
    \midrule
        \texttt{GPT-4} & \textbf{0.980}$\pm$\textbf{0.000} & \textbf{0.980}$\pm$\textbf{0.000} & \textbf{0.980}$\pm$\textbf{0.000} & 0.006$^*\pm$0.000 & \textbf{0.001}$\pm$\textbf{0.000} & \textbf{0.002}$^*\pm$\textbf{0.001} & 0.007$^*\pm$0.001 \\
    \bottomrule 
  \end{tabular}
  \caption{Overall performance (higher is better), bias along demographic dimensions (lower is better), and the associated bootstrapped standard error of the examined de-identification methods and \texttt{GPT-4}. We measure the bias with recall equality difference and bold \texttt{GPT-4}'s scores when they are among the best two in each column. In particular, \texttt{GPT-4} achieves the highest precision, recall, and F1 and beats the runner-up strongly. Moreover, the asterisk next to a bias score indicates a statistically significant difference in performance at an adjusted significance level ($5\%$ for gender, $0.833\%$ for race, $1.667\%$ for popularity and decade). \texttt{GPT-4} delivers the lowest bias among the non-rule-based methods along the dimensions of race and popularity.}
  \label{tab:overall_performance_gpt4}
  \vspace{-10pt}
\end{table*}

\begin{figure*}[t!]
    \centering
    \includegraphics[width=\textwidth]{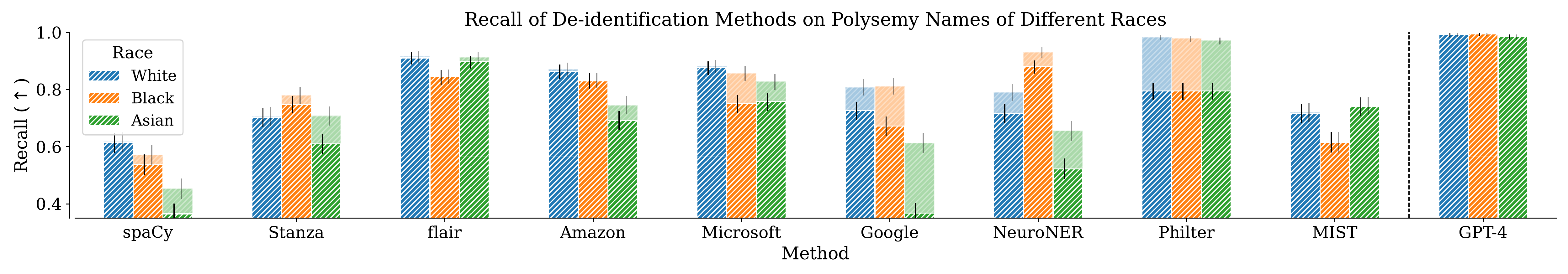}
    \vspace{-20pt}
    \caption{Recall and 95\% bootstrapped confidence interval on polysemy first names associated with three racial groups by each examined de-identification method and \texttt{GPT-4}. The increase in recall illustrated by the lighter color bar refers to the partially correct de-identification of non-polysemy last names. Unlike the other methods, \texttt{GPT-4} is robust to polysemy names across the three considered racial groups with almost no performance drop.}
    \label{fig:polysemy_recall_gpt4}
\end{figure*}

\section{GPT-4} \label{app:2}
To investigate the effectiveness of \texttt{GPT-4} \cite{OpenAI2023GPT4TR} in clinical record de-identification, we assess GPT-4-0613 via Azure OpenAI Service \cite{azureGPT} after opting out of human review of the data.
More specifically, we prompt \texttt{GPT-4} with a simple instruction (``Identify the names in the following clinical note. Output names only separated by commas.''), followed by each of the \numprint{16000} evaluation notes constructed in Section~\ref{sec:3}. 
We then locate the occurrences of the generated names by using pattern matching between \texttt{GPT-4}'s output and the corresponding input note.

\begin{figure*}[t!]
    \centering
    \includegraphics[width=\textwidth]{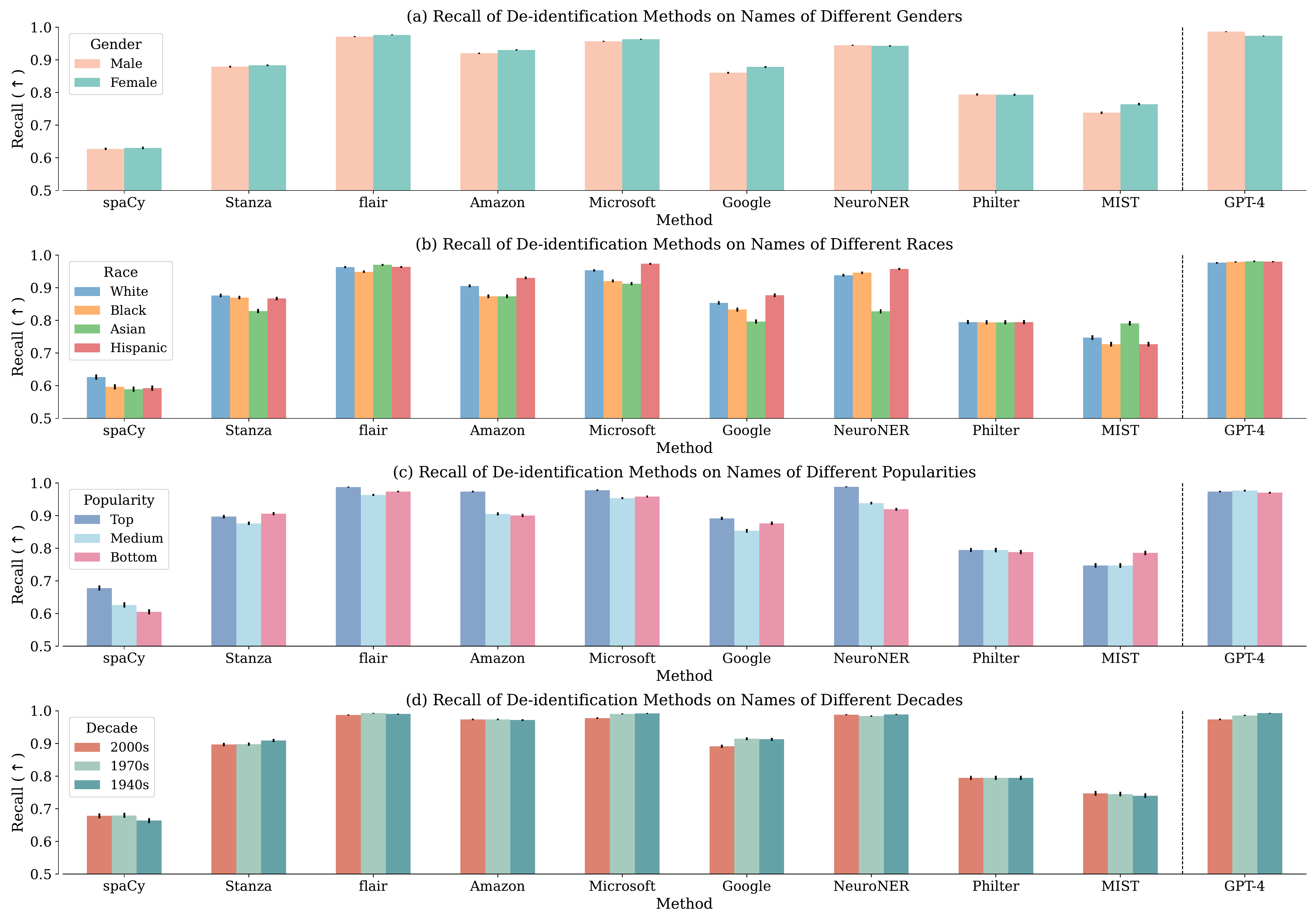}
    \caption{Recall and 95\% bootstrapped confidence interval of the demographic groups along each dimension by each examined de-identification method and \texttt{GPT-4}. Unlike the other methods, \texttt{GPT-4} demonstrates negligible disparities in performance along the dimensions of race and popularity but more significant performance gaps along the dimensions of gender and decade.}
    \label{fig:dimension_recall_gpt4}
\end{figure*}

The last row in Table~\ref{tab:overall_performance_gpt4} illustrates the overall performance of \texttt{GPT-4} and its bias along the four demographic dimensions. 
In particular, \texttt{GPT-4} is notably effective in de-identifying names in clinical notes, achieving the highest precision, recall, and F1 among the tested baselines and beating the second-best method (i.e., \texttt{flair}) by $0.024$ in F1. 
Furthermore, \texttt{GPT-4} also delivers the lowest bias (i.e., recall equality difference) among the non-rule-based de-identification methods along the dimensions of race and name popularity, where most other methods suffer. 
This result is further supported by the negligible gaps in recall among the racial and popularity groups considered in Figure~\ref{fig:dimension_recall_gpt4} (b) and (c), respectively. 
However, according to our hypothesis tests, \texttt{GPT-4} still acts differently with statistical significance along the dimensions of gender and decade of name popularity, which should be a call for further investigation for this commercial service.

\begin{figure*}[t!]
    \centering
    \includegraphics[width=0.75\textwidth]{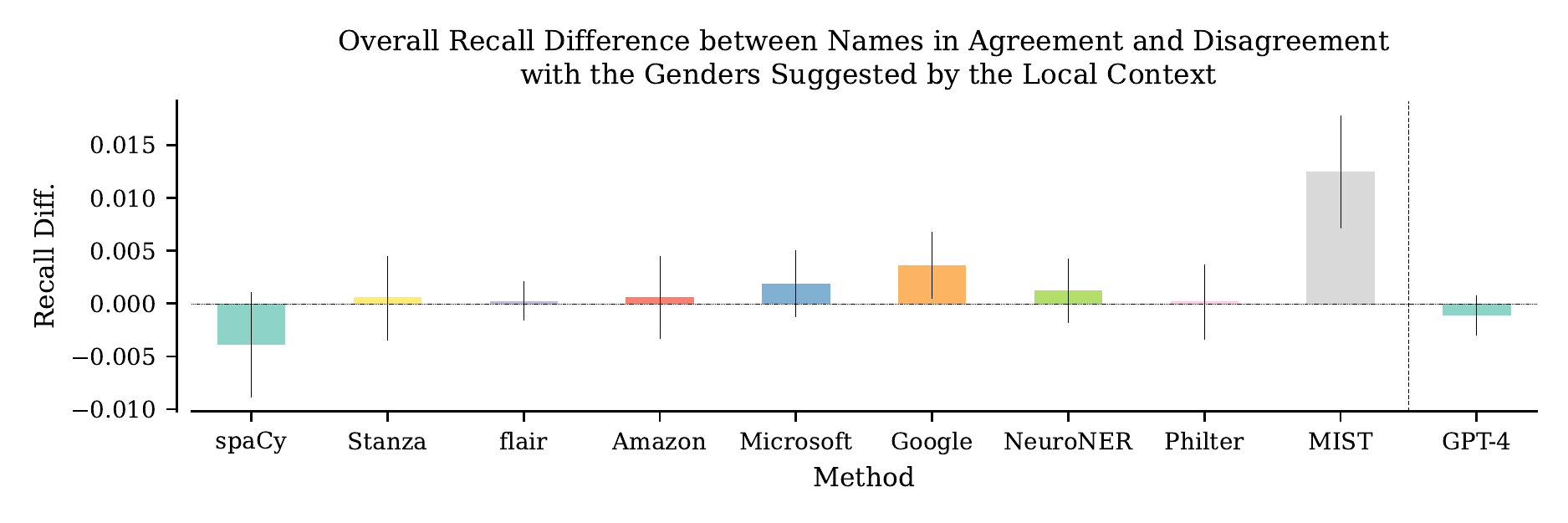}
    \caption{Difference in recall and 95\% bootstrapped confidence interval between names that are consistent and inconsistent with the genders suggested by the local context. A positive recall difference means that performance was best when there was gender consistency, while a negative recall difference means that performance was best when there was gender inconsistency. Notably, \texttt{GPT-4} performs better on names that are inconsistent with the gender suggested by the context.}
    \label{fig:context_difference_gpt4}
\end{figure*}

In addition, we examine the effect of polysemy names and context consistency on \texttt{GPT-4}'s performance in Figures~\ref{fig:polysemy_recall_gpt4} and~\ref{fig:context_difference_gpt4}, respectively.
Unlike most other tested methods, \texttt{GPT-4} is robust to polysemy names across the three considered racial groups with almost no performance drop, as shown in the rightmost column of Figure~\ref{fig:polysemy_recall_gpt4}.
On the other hand, similar to \texttt{spaCy}, \texttt{GPT-4} also performs better on names misaligned with the gender implied by the context, which suggests that \texttt{GPT-4} probably relies more on the memorized names, given its tremendous pre-training corpus, instead of the local context when inferring the occurrence of names.

These findings indicate that large language models (LLMs) like \texttt{GPT-4} can serve as an effective tool for de-identifying clinical notes \cite{agrawal2022large} but require further improvement in terms of mitigating the bias along the demographic dimensions of gender and decade of name popularity.
We leave the development of more advanced LLM-based de-identification methods to future work.
This research project has benefitted from the Microsoft Accelerate Foundation Models Research (AFMR) grant program through which leading foundation models hosted by Microsoft Azure, along with access to Azure credits, were provided to conduct the research.

\end{document}